\documentclass[10pt,twocolumn,letterpaper]{article}

\usepackage{wacv}
\usepackage{times}
\usepackage{epsfig}
\usepackage{graphicx}
\usepackage{amsmath}
\usepackage{amssymb}
\usepackage{subcaption}
\usepackage{enumitem}
\usepackage{tabularx}
\usepackage{algorithm}
\usepackage{algpseudocode}
\usepackage{url}
\usepackage{multirow}
\usepackage{rotating}
\usepackage{array}
\usepackage[table]{xcolor}
\usepackage{epstopdf}
\usepackage{booktabs}
\usepackage{amsthm}
\usepackage{caption}
\usepackage{scalerel}

\usepackage[pagebackref=true,breaklinks=true,letterpaper=true,colorlinks,bookmarks=false]{hyperref}

\DeclareMathOperator{\tr}{tr}
\newcommand{\PaperName}{IS4}

\newcommand{\sm}[2]{\scaleto{#1\mathstrut}{#2pt}}

\wacvfinalcopy 

\def\wacvPaperID{102} 

\ifwacvfinal\fi
\begin{document}

\title{Image Segmentation using Sparse Subset Selection}

\author{Fariba Zohrizadeh \hspace{1.2cm} Mohsen Kheirandishfard \hspace{1.2cm} Farhad Kamangar\\
Department of Computer Science and Engineering,\\
University of Texas at Arlington, USA\\
{\tt\small \{fariba.zohrizadeh,mohsen.kheirandishfard,farhad.kamangar\}@uta.edu}
}

\maketitle
\ifwacvfinal\thispagestyle{empty}\fi

\begin{abstract}
In this paper, we present a new image segmentation method based on the concept of sparse subset selection. Starting with an over-segmentation, we adopt local spectral histogram features to encode the visual information of the small segments into high-dimensional vectors, called superpixel features. Then, the superpixel features are fed into a novel convex model which efficiently leverages the features to group the superpixels into a proper number of coherent regions. Our model automatically determines the optimal number of coherent regions and superpixels assignment to shape final segments. To solve our model, we propose a numerical algorithm based on the alternating direction method of multipliers (ADMM), whose iterations consist of two highly parallelizable sub-problems. We show each sub-problem enjoys closed-form solution which makes the ADMM iterations computationally very efficient. Extensive experiments on benchmark image segmentation datasets demonstrate that our proposed method in combination with an over-segmentation can provide high quality and competitive results compared to the existing state-of-the-art methods.
\end{abstract}

\section{Introduction}
Image segmentation is a fundamental and challenging task in computer vision with diverse applications in various areas, such as video segmentation \cite{khoreva2014learning,khoreva2015classifier}, object segmentation \cite{dutt2016active,faridi2016automatic,ghiasi2016laplacian,jain2017fusionseg}, and semantic segmentation \cite{diaz2016lifting, krahenbuhl2011efficient}. The primary challenges of image segmentation are rooted in the diversity and ambiguity of visual textures encountered in input images. The solution to these challenges has been the subject of some research studies in the recent years \cite{arbelaez2011contour,fu2015robust,yuan2015factorization}.

\begin{figure}[t!]
\centering
	\begin{subfigure}[normal]{0.1\textwidth}
		\includegraphics[width=\textwidth]{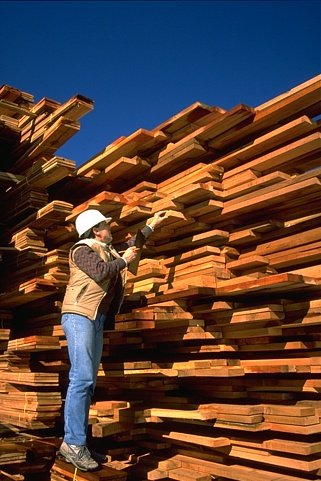}
		\caption{}
		\label{fig:spuriousClustersa}
	\end{subfigure}
	\begin{subfigure}[normal]{0.1\textwidth}
		\includegraphics[width=\textwidth]{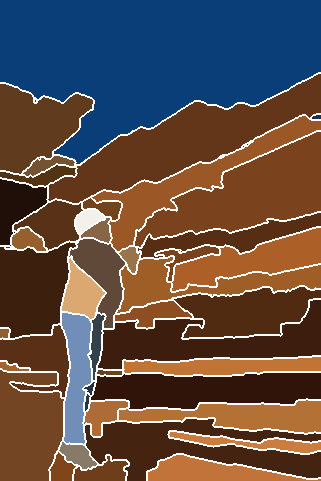}
		\caption{}
		\label{fig:spuriousClustersb}
	\end{subfigure}
	\begin{subfigure}[normal]{0.1\textwidth}
		\includegraphics[width=\textwidth]{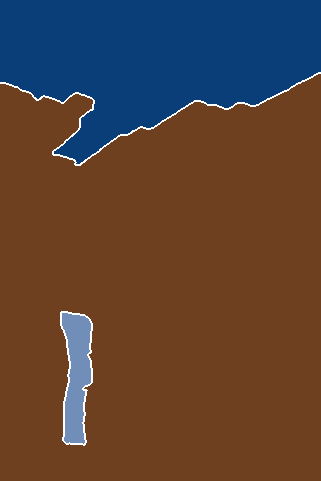}
		\caption{}
		\label{fig:spuriousClustersc}
	\end{subfigure}
	\begin{subfigure}[normal]{0.1\textwidth}
		\includegraphics[width=\textwidth]{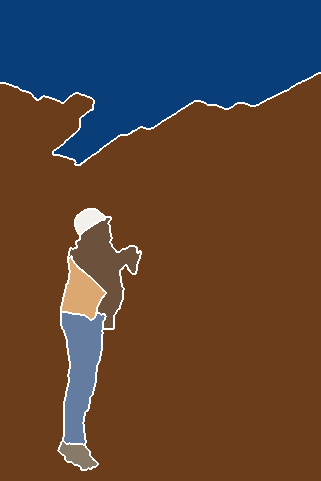}
		\caption{}
		\label{fig:spuriousClustersd}
	\end{subfigure}
\caption{Segmenting image pixels into multiple coherent regions. (a) Input image. (b-d) Segmentation results when the number of coherent regions is overestimated (b), underestimated (c), and properly determined (our method) (d).}
\label{fig:spuriousClusters}
\end{figure}

One of the major challenges in image segmentation is to determine the optimal number of coherent regions. This parameter can be calculated based on the distribution of image features \cite{yuan2015factorization}, given as a prior knowledge \cite{kim2013learning,li2012segmentation,zohrizadeh2016reliability} or set to a constant value \cite{arbelaez2011contour}, depending on the segmentation methodology. The performance of segmentation methods heavily depends on the right choice of this parameter, denoted by $K$. Figure \ref{fig:spuriousClusters} illustrates the segmentation results obtained for various choices of $K$. In the case that $K$ is overestimated (shown in Figure \ref{fig:spuriousClusters}\subref{fig:spuriousClustersb}), each coherent region may be divided into many separate segments. To merge these segments, many time-consuming and complicated steps are required which in turn increase the computational complexity and decrease the performance of the algorithm. On the other hand, when $K$ is underestimated (shown in Figure \ref{fig:spuriousClusters}\subref{fig:spuriousClustersc}), some coherent regions are forced to be merged together which in turn leads to a poor segmentation quality. Figure \ref{fig:spuriousClusters}\subref{fig:spuriousClustersd} shows our method has achieved a high quality segmentation by properly determining parameter $K$. 

Generally, determination of the number of coherent regions in image segmentation is nearly similar to the problem of finding the optimal number of clusters in other areas \cite{ben2001stability,still2004many,tibshirani2001estimating}. These problems are similar in a sense that they both seek to find the optimal number of groups. However, they may be different depending on the nature and intrinsic properties of their input data. For instance, in image segmentation, the spatial relationship among the image features can be perceived as an important property of data. Features may also have more specific properties depending on the feature extraction procedure. These properties need to be taken into account in determining the optimal number of coherent regions.

In this paper, we adopt local spectral histogram (LSH) features \cite{liu2006image} to model the input image. These features are computed by averaging the distribution of visual properties (such as color, texture, etc.) over a local patch centered at each pixel. Therefore, they can be considered as powerful tools to encode the local texture information. As LSH features are computed by averaging distributions in a local neighborhood, they are always nonnegative and the features belonging to the same coherent region are linearly dependent on each other. Our method leverages these properties to develop a convex model based on the concept of sparse subset selection. The main contributions of this work can be summarized as follows:

\noindent
\textbf{I}: We design an effective convex model based on the properties of LSH features which automatically determines the optimal number of coherent regions and pixels assignment.

\noindent
\textbf{II}: We develop a parallel numerical algorithm based on the alternating direction method of multiplier \cite{boyd2011distributed,gabay1976dual} whose iterations consists of two sub-problems with closed-form solutions. We show the proposed algorithm can solve our model significantly faster than the standard convex solvers \cite{mosek2015mosek,sturm1999using,toh1999sdpt3} while maintaining a high accuracy.

\noindent
\textbf{III}: We conduct extensive experiments on three commonly used datasets, BSD300 \cite{martin2001database}, BSD500 \cite{arbelaez2011contour}, and MSRC \cite{shotton2009textonboost} to show our results are competitive comparing to the results of state-of-the-arts methods.

The remainder of this paper is structured as follows: Section 2, shortly reviews related works; Section 3, explains our method in detail; Section 4, provides experimental results; Section 5, draws a conclusion about this paper.

\noindent
\textbf{Notation:} Throughout this paper, matrices, vectors, and scalars are denoted by boldface uppercase, boldface lowercase, and italic lowercase letters, respectively. For a given matrix $\mathbf{A}$, symbol ${\mathbf{A}}_{\sm{i,j}{6}}$ denotes the element at ${i}^{th}$ row and ${j}^{th}$ column, ${\left\|\mathbf{A}\right\|}_{\sm{F}{6}}$ indicates the Frobenius norm, and ${\left\|\mathbf{A}\right\|}_{\sm{p,q}{6}}$ is the ${\ell}_{\sm{p}{6}}$ norm of the ${\ell}_{\sm{q}{6}}$ norm of the rows in $\mathbf{A}$. For a given vector $\mathbf{a}$, symbols ${\left\|\mathbf{a}\right\|}_{\sm{p}{6}}$, $diag(\mathbf{a})$, and ${\mathrm{a}}_{\sm{i}{6}}$ denote standard ${\ell}_{\sm{p}{6}}$ norm, a diagonal matrix formed by the elements of $\mathbf{a}$, and the $i^{th}$ element of $\mathbf{a}$, respectively. Symbol $tr(.)$ stands for the trace operator, ${\mathbb{R}}_{\sm{+}{6}}$ indicates the set of positive real numbers, and $\textbf{1}$ is a column vector of all ones of appropriate dimension. 
\section{Related works}
Over-segmentation is obtained by partitioning the input image into multiple small homogeneous regions, called superpixels. Recent segmentation algorithms usually utilize an over-segmentation and merge the similar superpixels to shape final segments \cite{fu2015robust,li2012segmentation,zohrizadeh2016reliability}. Fu \etal \cite{fu2015robust} proposed a pipeline of three effective post-processing steps which are applied on an over-segmentation to shape final segments. Li \etal \cite{li2012segmentation} suggested to construct a bipartite graph over multiple over-segmentations provided by \cite{comaniciu2002mean} and \cite{felzenszwalb2004efficient}. Then, the spectral clustering is applied on the graph to form final segments. Ren \etal \cite{ren2013image} presented a method which constructs a cascade of boundary classifiers and iteratively merges the superpixels of an over-segmentation to build final segments.

One notable segmentation method is presented by Arbelaez \etal in \cite{arbelaez2011contour}, which reduces the problem of image segmentation to a contour detection. The method combines multiple contour cues of different image layers to create a contour map, called gPb. The contour map and its corresponding hierarchical segmentation further utilized by some algorithms as an initial over-segmentation \cite{donoser2014discrete,gao2016graph,kim2011boundary,7515198}. Liu \etal \cite{7515198} trained a classifier over the gPb results to construct a hierarchical region merging tree. Then, the classifier iteratively merges the most similar regions to shape final segments. Gao \etal \cite{gao2016graph} proposed to construct a graph over the gPb results based on the spatial and visual information of superpixels. Then, a model is proposed to partition the graph into multiple components where each one is corresponding to a final segment. Recently, a widely-used extension of gPb is proposed by Arbelaez \etal \cite{Arbelaez_2014_CVPR}, called Multiscale Combinatorial Grouping (MCG). The method combines the information obtained from multiple image scales to generate a hierarchical segmentation. Yu \etal \cite{yu2015piecewise} proposed a nonlinear embedding method based on a $\ell_{\sm{1}{6}}$-regularized objective which is integrated into MCG framework to provide better local distances among the superpixels. Chen \etal \cite{chen2016scale} realigned the MCG results by modifying the depth of segments in its hierarchical structure. 
\begin{figure*}[t!]
\centering
		\includegraphics[width=0.9\textwidth]{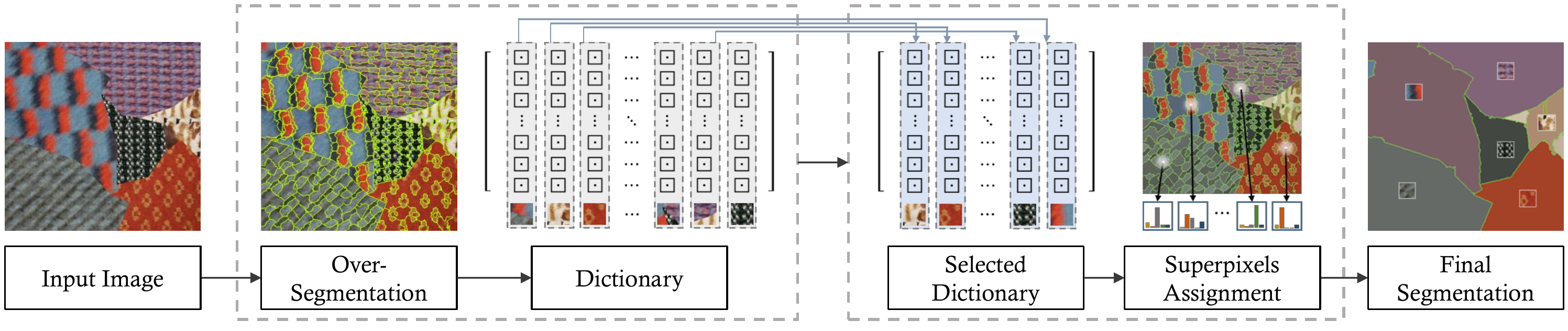}
		\caption{The pipeline of our proposed algorithm. Given an input image, we adopt an algorithm to generate a super-pixel segmentation layer. Then, we compute the superpixels features and learn a dictionary of words over all superpixels. Our convex model efficiently selects a small subset of informative words and softly assigns superpixels to the selected words. The neighboring superpixels which are assigned to the same selected words are merged to shape final segmentation.}
	\label{fig:graph}
\end{figure*}
There are some recently-developed methods based on deep learning in many related tasks such as contour detection \cite{kokkinos2015pushing,Maninis2017} and semantic image segmentation \cite{hariharan2014simultaneous,long2015fully,noh2015learning,zheng2015conditional}. Although these methods are able to exploit more sophisticated and complex representative features, they are often highly demanding in terms of training data and training time. Therefore, these methods may not be the most appropriate choice in some applications. To illustrate, consider natural image segmentation in which many unknown and diverse patterns are likely to be presented in each single image. This implies that we may have insufficient number of training samples per each pattern. Motivated by this, we propose a new image segmentation method based on the concept of sparse subset selection \cite{7364258,esser2012convex}. The method starts with an over-segmentation (e.g., MCG) and uses an effective convex model to group the superpixels into a proper number of coherent regions. Our work is roughly similar to the factorization-based segmentation (Fact) algorithm \cite{yuan2015factorization} in the sense that both use local spectral histogram (LSH) features to model the input image and seek to estimate the optimal number of coherent regions. However, our method differs from Fact in two major ways: (1) Fact determines the optimal number of coherent regions and pixels assignment in two consecutive steps which may lead to error propagation, but our model simultaneously determines the optimal number of coherent regions and pixels assignment in an effective manner. (2) Fact does not take advantage of the spatial information among pixels, but we incorporate this information as a Laplacian regularization term in our convex model. Moreover, we propose a parallel numerical algorithm based on the alternating direction method of multipliers \cite{boyd2011distributed,gabay1976dual} to solve our model and obtain final segments. Note that the model can be easily utilized in some other applications such as video summarization \cite{7364258,elhamifar2012see} and dimensionality reduction \cite{esser2012convex}.

\section{Proposed method} \label{sec:PropMe}
This section describes our segmentation method in two phases: problem formulation and numerical algorithm. The first phase formulates a convex model based on the properties of local spectral histogram (LSH) features and the second phase presents our solution to the model in details. 

\subsection{Problem formulation}
Given an input image $\mathbf{I}$, we start with an over-segmentation consisting $n$ superpixels. We form feature matrix $\mathbf{X}=\left[{\mathbf{x}}_{\sm{1}{6}}|{\mathbf{x}}_{\sm{2}{6}}|\dots|{\mathbf{x}}_{\sm{n}{6}}\right]\in{\mathbb{R}}_{{\sm{+}{6}}}^{{\sm{d\times n}{6}}}$ by averaging the LSH features of pixels within each superpixel. Hence, each ${\mathbf{x}}_{\sm{i}{6}}$ is considered as a $d$-dimensional feature corresponding to the $i^{th}$ superpixel. Under the assumption of linear dependence among the LSH features, we model the feature matrix $\mathbf{X}$ as,
\begin{equation} \label{eq:imageModel}
\begin{aligned}
\mathbf{X}=\mathbf{D} \mathbf{U}+\mathbf{E},
\end{aligned}
\end{equation}
where $\mathbf{D}\!=\![{\mathbf{d}}_{\sm{1}{6}}|{\mathbf{d}}_{\sm{2}{6}}|\dots|{\mathbf{d}}_{\sm{l}{6}}]\!\in\!{\mathbb{R}}_{\sm{+}{6}}^{\sm{d\times l}{6}}$ is a dictionary of $l$ words inferred from the superpixels features, $\mathbf{U}\!=\!{[{\mathbf{u}}_{\sm{1}{6}}|{\mathbf{u}}_{\sm{2}{6}}|\dots|{\mathbf{u}}_{\sm{n}{6}}]}\in {\mathbb{R}}^{\sm{l\times n}{6}}$ denotes a coefficient matrix whose rows indicate the contribution of each word in reconstructing $\mathbf{X}$, and $\mathbf{E}\!\in\!{\mathbb{R}}^{\sm{d\times n}{6}}$ indicates the model error. The goal is to design a model which takes into account the linear dependence and spatial relationship among the features to compute an optimal matrix $\mathbf{U}$.

In order to incorporate the linear dependence among the features into the model, we adopt a non-negative matrix factorization framework to construct $\mathbf{D}$ over the feature matrix $\mathbf{X}$. Consider the dissimilarity matrix $\mathbf{R}\in\mathbb{R}^{\sm{l\times n}{6}}$ where ${\mathbf{R}}_{\sm{j,i}{6}}$ indicates the dissimilarity between ${\mathbf{d}}_{\sm{j}{6}}$ and ${\mathbf{x}}_{\sm{i}{6}}$. We use ${\mathbf{R}}_{\sm{j,i}{6}}={\left\|{\mathbf{d}}_{\sm{j}{6}}-{\mathbf{x}}_{\sm{i}{6}}\right\|}_{\sm{2}{6}}^{\sm{2}{6}}$ to compute $\mathbf{R}$. In the case of normalized features and visual words, ${\mathbf{R}}_{\sm{j,i}{6}}$ only depends on the inner product between ${{\mathbf{d}}_{\sm{j}{6}}}$ and ${\mathbf{x}}_{\sm{i}{6}}$. More precisely, it shows how well ${\mathbf{x}}_{\sm{i}{6}}$ is expressible by ${\mathbf{d}}_{\sm{j}{6}}$ which is a reasonable dissimilarity measure according to the linear dependence among the features. Since the superpixels are not necessarily of the same size, we define a diagonal regularization matrix $\mathbf{P}\in\mathbb{R}^{\sm{n\times n}{6}}$ whose diagonal elements are proportional to the number of pixels in each superpixel. The elements of $\mathbf{P}$ scale each ${\mathbf{R}}_{\sm{j,i}{6}}$ by the size of the $i^{th}$ superpixel.

In order to embed the spatial relationship among the superpixels into the model, we construct a graph over the initial over-segmentation. Let $\mathcal{G}=(\mathcal{V},\mathcal{E},\mathbf{W})$ be the graph where nodes are superpixels and edges connect every pairs of adjacent superpixels with a weight specified by $\mathbf{W}\in\mathbb{R}^{\sm{n\times n}{6}}$. The edge weight between the adjacent superpixels $i$ and $j$ indicates their similarity and is defined as:
\begin{equation} \label{eq:edgeWeights}
\begin{aligned}
& {\mathbf{W}}_{i,j}=e^{-\frac{{{\left\|{\mathbf{x}}_i-{\mathbf{x}}_j\right\|}}_2^2}{{\sigma}_x}-\overline{b}},
\end{aligned}
\end{equation}
where $\overline{b}$ is the average strength of their common boundary and ${\sigma}_{\sm{x}{6}}$ controls the effect of feature distances on their similarity weight. Given such graph $\mathcal{G}$, we define Laplacian matrix $\mathbf{L}\in\mathbb{R}^{\sm{n\times n}{6}}$ as $\mathbf{L}=diag(\mathbf{W}\mathbf{1})-\mathbf{W}$. 

Once the Laplacian matrix $\mathbf{L}$, the dissimilarity matrix $\mathbf{R}$, and the regularization matrix $\mathbf{P}$ are computed, we seek to find a small subset of the dictionary words that well represents feature matrix $\mathbf{X}$. To do so, a model is required which satisfies the following requirements:
\begin{itemize}
\itemsep0em 
\item minimizes the number of selected words. In the ideal case, we are interested to have a single word corresponding to each coherent region. 
\item ensures each feature in ${\{{\mathbf{x}}_{\sm{i}{6}}\}}_{\sm{i=1}{6}}^{\sm{n}{6}}$ is well expressible as a nonnegative linear combination of the selected words. The coefficients of such linear combination indicate the contribution of each selected word in reconstructing the feature.
\item ensures each feature in ${\{{\mathbf{x}}_{\sm{i}{6}}\}}_{\sm{i=1}{6}}^{\sm{n}{6}}$ is expressed by at least one selected word. To do so, we impose a constraint on the sum of the linear combination coefficients.
\item takes advantage of the spatial relationship and linear dependence of the features.
\end{itemize}
Motivated by \cite{7364258}, we formulate the following convex model which fulfills the requirements.
\begin{subequations} \label{eq:OurModel}
\begin{align}
\underset{
\mathbf{U}\in{\mathbb{R}}^{l\times n}
}{\text{minimize}}
&\;\tr(\mathbf{P}\mathbf{R}\!^{\top}\mathbf{U})\!+\!\gamma \tr (\mathbf{U} \mathbf{L} {\mathbf{U}}^{\top}\!)
\!+\!\lambda {\left\|\mathbf{U}\right\|}_{1,\infty} \label{eq:OurModels1}\\
\text{subject to}
&\; \mathbf{U}\geq 0, \label{eq:OurModels2} \\
&\; {\textbf{1}}^{\top} \mathbf{U} = {\textbf{1}}^{\top}\!, \label{eq:OurModels3}
\end{align}
\end{subequations}
where $\gamma>0$ and $\lambda>0$ are regularization parameters. The first term in \eqref{eq:OurModel} is corresponding to the cost of representing feature matrix $\mathbf{X}$ using dictionary $\mathbf{D}$ proportional to the size of superpixels. The Laplacian regularization term incorporates the spatial relation of superpixels into the objective and the last term is a row sparsity regularization term which penalizes the objective in proportion to the number of selected words. Note that although $\mathbf{D}$ does not directly appear in \eqref{eq:OurModel}, the rows of $\mathbf{U}$ are constructed based on the contribution of the dictionary words, $\{{\mathbf{d}}_{\sm{j}{6}}\}_{\sm{j=1}{6}}^{\sm{l}{6}}$, in reconstructing $\mathbf{X}$. 

The optimal solution of problem \eqref{eq:OurModel} is ${\mathbf{U}}^{\sm{\ast}{6}}\in {\left[0,1\right]}^{\sm{l\times n}{6}}$ whose nonzero rows are corresponding to the selected words. Note that ${\mathbf{U}}^{\sm{\ast}{6}}$ not only determines the selected words but also shows the contribution of selected words in reconstructing the superpixel features $\{{\mathbf{x}}_{\sm{i}{6}}\}_{\sm{i=1}{6}}^{\sm{n}{6}}$. Hence, the elements of ${\mathbf{U}}^{\sm{\ast}{6}}$ can be interpreted as a soft assignment of the superpixels to the selected words. In this case, the $i^{th}$ superpixel is assigned to the selected word which has the largest contribution in the reconstruction of ${\mathbf{x}}_{\sm{i}{6}}$. Final segmentation is obtained by merging the neighboring superpixels which are assigned to the same selected word. Figure \ref{fig:graph} illustrates our segmentation pipeline in details.

\subsection{Numerical algorithm}\label{sec:NumAlg}
To solve the model, we propose an efficient numerical algorithm based on alternating direction method of multipliers (ADMM) which is a powerful technique to solve convex optimization problems. The basic idea of ADMM is to introduce auxiliary variables to break down a complicated convex problem into smaller sub-problems where each one is efficiently solvable via an explicit formulas. The ADMM iteratively solves the sub-problems until convergence. To formulate the ADMM, let define $\mathbf{m}\in\mathbb{R}^{\sm{l}{6}}$ such that $\mathrm{m}_{\sm{j}{6}}=\text{argmax}_{\sm{i}{6}}\:{|\mathbf{U}_{\sm{j,i}{6}}|}$ and reformulate \eqref{eq:OurModel} as follows:
\begin{subequations} \label{eq:ADMMEq1}
\begin{align}
\underset{\begin{subarray}{l}
\mathbf{U}\in\mathbb{R}^{l\times n}\\\mathbf{m}\in\mathbb{R}^{l}
\end{subarray}}
{\text{minimize}}
&\;\tr(\mathbf{P}\mathbf{R}\!^{\top}\mathbf{U})\!+\!\gamma \tr (\mathbf{U} \mathbf{L} {\mathbf{U}}^{\top}\!)\!+\!\lambda {\mathbf{1}}^{\top}\!\mathbf{m} \label{eq:ADMMEq1s1}\\
\text{subject to}
&\; \mathbf{U}\geq 0, \label{eq:ADMMEq1s2}\\
&\; {\textbf{1}}^{\top} \mathbf{U} = {\textbf{1}}^{\top}\!, \label{eq:ADMMEq1s3} \\
&\; \mathbf{m}{\mathbf{1}}^{\top}\geq \mathbf{U}, \label{eq:ADMMEq1s4}
\end{align}
\end{subequations}
where \eqref{eq:ADMMEq1s4} is imposed to guarantee the equivalence of \eqref{eq:OurModel} and \eqref{eq:ADMMEq1}. Using slack variable $\mathbf{V}={[{\mathbf{v}}_{\sm{1}{6}}|\dots|{\mathbf{v}}_{\sm{n}{6}}]}\in\mathbb{R}^{\sm{l\times n}{6}}$, \eqref{eq:ADMMEq1} can be written in more convenient form as,
\begin{subequations} \label{eq:ADMMEq2}
\begin{align}
\underset{\begin{subarray}{l}
\mathbf{U},\mathbf{V}\in\mathbb{R}^{l\times n}\\\mathbf{m}\in\mathbb{R}^{l}
\end{subarray}}
{\text{minimize}}
&\;\tr(\mathbf{P}\mathbf{R}\!^{\top}\mathbf{U})\!+\!\gamma \tr (\mathbf{U} \mathbf{L} {\mathbf{U}}^{\top}\!)\!+\!\lambda {\mathbf{1}}^{\top}\!\mathbf{m} \label{eq:ADMMEq2s1}\\
\text{subject to}
&\;\mathbf{U}\geq 0, \label{eq:ADMMEq2s2} \\
&\; {\textbf{1}}^{\top} \mathbf{U} = {\textbf{1}}^{\top}\!, \label{eq:ADMMEq2s3} \\
&\; \mathbf{m}{\mathbf{1}}^{\top}=\mathbf{V}+\mathbf{U}, \label{eq:ADMMEq2s4} \\
&\; \mathbf{V}\geq 0. \label{eq:ADMMEq2s5}
\end{align}
\end{subequations}

As ${\mathbf{1}}^{\top}\mathbf{m}{\mathbf{1}}^{\top}\mathbf{1}\!=\!{\mathbf{1}}^{\top}(\mathbf{V}+\mathbf{U})\mathbf{1}$, the third term of \eqref{eq:ADMMEq2s1} can be equivalently written as $\frac{\lambda}{n}{\mathbf{1}}^{\top}(\mathbf{V}+\mathbf{U})\mathbf{1}$. Hence, \eqref{eq:ADMMEq2} can be reformulated independent of $\mathbf{m}$ as:  
\begin{subequations} \label{eq:ADMMEq3}
\begin{align}
\underset{\begin{subarray}{l}
\mathbf{U},\mathbf{V}\in\mathbb{R}^{l\times n}
\end{subarray}}
{\text{minimize}}
& \tr(\mathbf{P}\mathbf{R}\!^{\top}\!\mathbf{U})\!+\!\gamma \tr (\mathbf{U} \mathbf{L} {\mathbf{U}}^{\top}\!)\!+\!\frac{\lambda}{n}{\mathbf{1}}\!^{\top}\!(\mathbf{V}\!+\!\mathbf{U})\mathbf{1} \!\!\!\!\label{eq:ADMMEq3s1}\\
\text{subject to}
&\; \mathbf{U}\geq 0, \label{eq:ADMMEq3s2} \\
&\; {\textbf{1}}^{\top} \mathbf{U} = {\textbf{1}}^{\top}\!, \label{eq:ADMMEq3s3} \\
&\; {\mathbf{v}}_{i-1}+{\mathbf{u}}_{i-1}={\mathbf{v}}_{i}+{\mathbf{u}}_{i},\;\; \sm{i=2,\dots,n}{8},\label{eq:ADMMEq3s4} \\
&\; \mathbf{V}\geq 0, \label{eq:ADMMEq3s5}
\end{align}
\end{subequations}
where \eqref{eq:ADMMEq3s4} is obtained by removing  $\mathbf{m}$ from \eqref{eq:ADMMEq2s4}. In order to derive an ADMM formulation with subproblems possessing explicit formulas, we introduce auxiliary matrices $\hat{\mathbf{U}}\in\mathbb{R}^{\sm{l\times n}{6}},\hat{\mathbf{V}}\in\mathbb{R}^{\sm{l\times n}{6}}$ and reformulate \eqref{eq:ADMMEq3} as:

\begin{subequations} \label{eq:ADMMEq4}
\begin{align}
\underset{\begin{subarray}{l}
\mathbf{U},\!\mathbf{V}\!,\hat{\mathbf{U}},\!\hat{\mathbf{V}}\!\in\mathbb{R}^{l\times n}
\end{subarray}}
{\text{minimize}}
&\tr(\mathbf{P}\mathbf{R}\!^{\top}\mathbf{\hat{U}})\!+\!\gamma \tr (\mathbf{U} \mathbf{L} {\mathbf{U}}^{\top}\!)\!+\!\frac{\lambda}{n}{\mathbf{1}}\!^{\top}\!(\mathbf{V}\!\!+\!\mathbf{U})\mathbf{1} \nonumber\\
&+\frac{{\mu}_1}{2}{\left\|\mathbf{U}-\hat{\mathbf{U}}\right\|}_F^2+\frac{{\mu}_2}{2}{\left\|\mathbf{V}-\hat{\mathbf{V}}\right\|}_F^2 \label{eq:ADMMEq4s1}\\
\text{subject to}
&\; \mathbf{\hat{U}}\geq 0, \label{eq:ADMMEq4s2} \\
&\; {\textbf{1}}^{\top} \mathbf{\hat{U}}={\textbf{1}}^{\top}\!, \label{eq:ADMMEq4s3} \\
&\; {\mathbf{v}}_{i-1}+{\mathbf{u}}_{i-1}={\mathbf{v}}_{i}+{\mathbf{u}}_{i},\;\; \sm{i=2,\dots,n}{8},\label{eq:ADMMEq4s4} \\
&\; \mathbf{\hat{V}}\geq 0, \label{eq:ADMMEq4s5} \\
&\; \mathbf{U}=\mathbf{\hat{U}}, \label{eq:ADMMEq4s6} \\
&\; \mathbf{V}=\mathbf{\hat{V}}, \label{eq:ADMMEq4s7}
\end{align}
\end{subequations}
where ${\mu}_{\sm{1}{6}}\!>\!0$ and ${\mu}_{\sm{2}{6}}\!>\!0$ are the augmented Lagrangian parameters. As it is suggested in \cite{boyd2011distributed}, we can set ${\mu}_{\sm{1}{6}}\!=\!{\mu}_{\sm{2}{6}}\!=\!\mu$. Note that \eqref{eq:ADMMEq4} is equivalent to \eqref{eq:ADMMEq3}, because the additional terms in \eqref{eq:ADMMEq4s1} vanish for any feasible solution. To solve \eqref{eq:ADMMEq4}, augmented Lagrangian function is formed as:
\begin{equation} \label{eq:ADMMEq5}
\begin{aligned}\!\!
&\!\!\!\mathcal{L}_{\mu}(\mathbf{U},\!\mathbf{V},\!\hat{\mathbf{U}},\!\hat{\mathbf{V}},\!{\mathbf{\Lambda}}_1,\!{\mathbf{\Lambda}}_2)\!=\!\tr(\mathbf{P}\mathbf{R}\!^{\top}\mathbf{\hat{U}})\!+\!\gamma \tr (\mathbf{U} \mathbf{L} {\mathbf{U}}^{\top}\!)\\
&\!\!\!\!\!+\!\frac{\lambda}{n}{\mathbf{1}}\!^{\top}\!\!(\mathbf{V}\!\!+\!\mathbf{U})\mathbf{1}\!+\!\frac{{\mu}}{2}{\left\|\mathbf{U}\!-\!\hat{\mathbf{U}}\!+\!\frac{{\mathbf{\Lambda}}_1}{{\mu}}\right\|}_F^2\!\!\!\!+\!\frac{{\mu}}{2}{\left\|\mathbf{V}\!\!-\!\!\hat{\mathbf{V}}\!\!+\!\frac{{\mathbf{\Lambda}}_2}{{\mu}}\right\|}_F^2 
\end{aligned}\!\!\!
\end{equation}
where ${\mathbf{\Lambda}}_{\sm{1}{6}}\in\mathbb{R}^{\sm{l\times n}{6}}$ and ${\mathbf{\Lambda}}_{\sm{2}{6}}\in\mathbb{R}^{\sm{l\times n}{6}}$ are Lagrange multipliers associated with the equality constraints \eqref{eq:ADMMEq4s6} and \eqref{eq:ADMMEq4s7}. 

Given initial values for $\hat{\mathbf{U}}$, $\hat{\mathbf{V}}$, ${\mathbf{\Lambda}}_{\sm{1}{6}}$, and ${\mathbf{\Lambda}}_{\sm{2}{6}}$, the ADMM iterations to solve \eqref{eq:ADMMEq4} are summarized as follow:
\begin{equation} \label{eq:ADMMBlock1}
\begin{aligned}
\hspace{-6mm}({\mathbf{U}}^{k+1}\!,\!\!{\mathbf{V}}^{k+1})\!\!:=\!\!& \underset{\begin{subarray}{l}
{\mathbf{U}}\!,\mathbf{V}\in \mathbb{R}^{l\times n}
\end{subarray}}
{\text{argmin}}\!\!\!\!\!\!\!\!
& &\!\mathcal{L}_{\mu}(\mathbf{U},\!\mathbf{V}\!,{\hat{\mathbf{U}}}^k,\!{\hat{\mathbf{V}}}^k,{\mathbf{\Lambda}}_1^k,{\mathbf{\Lambda}}_2^k) \\
\hspace{-6mm}& \text{subject to}\hspace{-3mm}
& &{\mathbf{v}}_{i-1}\!+\!{\mathbf{u}}_{i-1}\!=\!{\mathbf{v}}_{i}\!+\!{\mathbf{u}}_{i},\hspace{1mm}\sm{i=2,\dots,n}{8},
\end{aligned}\!\!\!\!\!\!\!\!\!\!\!
\end{equation}

\begin{equation} \label{eq:ADMMBlock2}
\begin{aligned}
\hspace{-6mm}({\hat{\mathbf{U}}}^{k+1}\!,\!\!{\hat{\mathbf{V}}}^{k+1})\!\!:=\!\!& \underset{\begin{subarray}{l}
{\hat{\mathbf{U}}}\!,{\hat{\mathbf{V}}}\in \mathbb{R}^{l\times n}
\end{subarray}}
{\text{argmin}}\!\!\!\!\!\!\!\!
& &\!\!\!\mathcal{L}_{\mu}({\mathbf{U}}^{k+1}\!,\!{\mathbf{V}}^{k+1}\!,\!{\hat{\mathbf{U}}}\!,\!{\hat{\mathbf{V}}}\!,{\mathbf{\Lambda}}_1^k\!,{\mathbf{\Lambda}}_2^k) \\
\hspace{-6mm}& \text{subject to}\hspace{-3mm}
& & \hat{\mathbf{U}}\geq 0,\;{\textbf{1}}^{\top} \mathbf{\hat{U}}={\textbf{1}}^{\top}\!, \\
&&& \hat{\mathbf{V}}\geq 0.
\end{aligned}\!\!\!\!\!\!\!\!\!\!\!
\end{equation}

\begin{equation} \label{eq:ADMMBlock3}
\begin{aligned}
\begin{cases}
& {\mathbf{\Lambda}}_1^{k+1}={\mathbf{\Lambda}}_1^{k}+{\mu}({\mathbf{U}}^{k+1}-{\hat{\mathbf{U}}}^{k+1}) \\
& {\mathbf{\Lambda}}_2^{k+1}={\mathbf{\Lambda}}_2^{k}+{\mu}({\mathbf{V}}^{k+1}-{\hat{\mathbf{V}}}^{k+1})
\end{cases}
\end{aligned}
\end{equation}
Note that the variables in each of the above iterations can be stacked together to form a single matrix variable. Therefore, the numerical algorithm is not considered as a multi-block ADMM. To solve \eqref{eq:ADMMBlock1}, let form ${\mathbf{y}}_{\sm{j}{6}}\in \mathbb{R}^{\sm{2n}{6}}$ by concatenating the $j^{th}$ rows of $\mathbf{U}$ and $\mathbf{V}$. Then, \eqref{eq:ADMMBlock1} can be divided into $l$ equality constrained quadratic programs as follows:
\begin{subequations} \label{eq:ADMMEq6}
\begin{align}
\underset{\begin{subarray}{l}
{\mathbf{y}}_j\in \mathbb{R}^{2n}
\end{subarray}}
{\text{minimize}}
&\;\frac{1}{2}{{\mathbf{y}}_j}^{\top}\!\mathbf{B}{\mathbf{y}}_j+{{\mathbf{y}}_j}^{\top}\!{\mathbf{b}}_j\label{eq:ADMMEq6s1}\\
\text{subject to}
&\; \mathbf{A}{\mathbf{y}}_j={\mathbf{c}}, \label{eq:ADMMEq6s2} 
\end{align}
\end{subequations}
where $\mathbf{B}\in\mathbb{R}^{\sm{2n\times 2n}{6}}$ is a block diagonal positive semi-definite matrix, $\mathbf{A}\in\mathbb{R}^{\sm{n\times 2n}{6}}$ is a sparse matrix corresponding to the constraint \eqref{eq:ADMMEq4s4}, and $\mathbf{c}\!\in\!\mathbb{R}^{\sm{n}{6}}$ is a vector of all zeros.

Problem \eqref{eq:ADMMBlock2} can be split into two separate sub-problems with closed-form solutions as follows:
\begin{subequations} \label{eq:ADMMEq7}
\begin{align}
& \underset{\begin{subarray}{l}
\hat{\mathbf{U}}\in{\mathbb{R}}^{l\times n}
\end{subarray}}
{\text{minimize}}
& & {\left\|\hat{\mathbf{U}}-(\mathbf{U}+\frac{{\mathbf{\Lambda}}_1+\mathbf{R}\mathbf{P}\!^{\top}}{\mu})\right\|}_F^2 \label{eq:ADMMEq7s1}\\
& \text{subject to}
& &\mathbf{\hat{U}}\geq 0, {\textbf{1}}^{\top} \mathbf{\hat{U}}={\textbf{1}}^{\top}, \label{eq:ADMMEq7s2} 
\end{align}
\end{subequations}
\begin{subequations} \label{eq:ADMMEq8}
\begin{align}
& \underset{\begin{subarray}{l}
\hat{\mathbf{V}}\in{\mathbb{R}}^{l\times n}
\end{subarray}}
{\text{minimize}}
& & {\left\|\hat{\mathbf{V}}-(\mathbf{V}+\frac{{\mathbf{\Lambda}}_2}{\mu})\right\|}_F^2  \label{eq:ADMMEq8s1}\\
& \text{subject to}
& &\mathbf{\hat{V}}\geq 0, \label{eq:ADMMEq8s2} 
\end{align}
\end{subequations}
where each one consists of $n$ computationally cheap parallel programs. Sub-problem \eqref{eq:ADMMEq7} can be divided into $n$ parallel programs over the columns of $\hat{\mathbf{U}}$ where each one is a Euclidean norm projection onto the probability simplex constraints. These programs enjoy closed-form solutions as presented in \cite{wang2013projection}. Sub-problem \eqref{eq:ADMMEq8} consists of $n$ small parallel programs over the columns of $\hat{\mathbf{V}}$, where each program is a minimization of the Euclidean norm projection onto the nonnegative orthant and admits closed-form solution.

Problem \eqref{eq:ADMMBlock3} can be split into two sub-problems over ${\mathbf{\Lambda}}_{\sm{1}{6}}$ and ${\mathbf{\Lambda}}_{\sm{2}{6}}$ where each sub-problem consists of $n$ parallel updates over the columns of corresponding matrix.

\begin{figure}[t!]
\centering
\includegraphics[width=\linewidth]{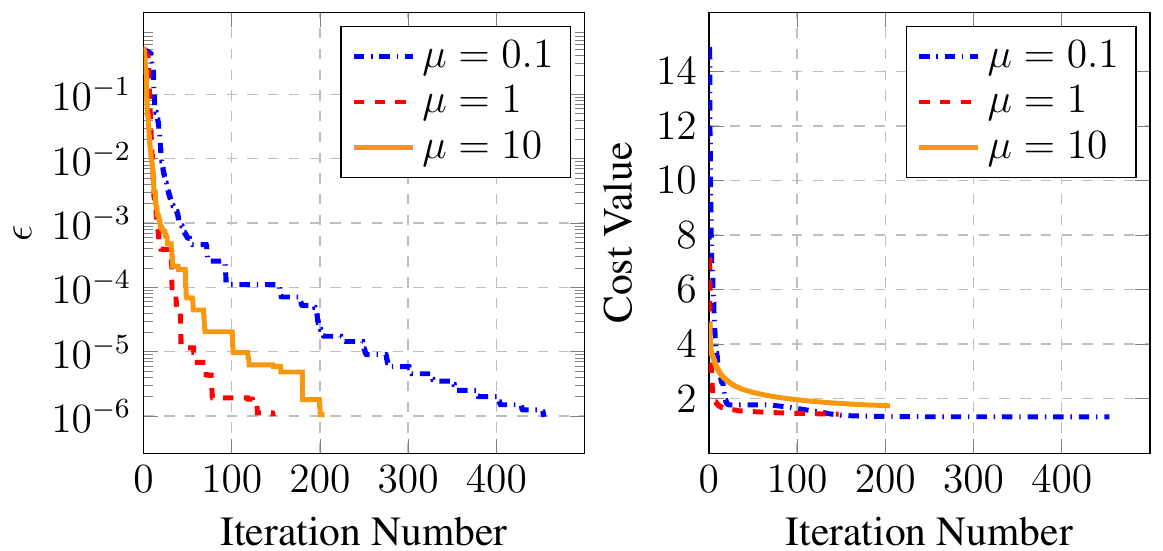}
\caption{Our convergence behavior for different choices of ${\mu}$. Left: Combined residual. Right: Cost function.}
\label{fig:ConvBehav}
\end{figure}

Our numerical algorithm consists of two sub-problems with closed-form solutions, which makes the iterations computationally efficient. To evaluate the convergence behavior of the proposed algorithm, we adopt combined residual presented in \cite{goldstein2014fast} as:
\begin{equation}
\begin{aligned}
\mathbf{\epsilon}^{k+1}&=\frac{1}{\mu} {\left\|{\mathbf{\Lambda}}_1^{k+1}-{\mathbf{\Lambda}}_1^k\right\|}_F^2+{\mu} {\left\|{\mathbf{U}}^{k+1}-{\mathbf{U}}^k\right\|}_F^2 \\ 
& +\frac{1}{\mu} {\left\|{\mathbf{\Lambda}}_2^{k+1}-{\mathbf{\Lambda}}_2^k\right\|}_F^2+{\mu} {\left\|{\mathbf{V}}^{k+1}-{\mathbf{V}}^k\right\|}_F^2.
\end{aligned}
\end{equation}
Figure \ref{fig:ConvBehav} demonstrates the convergence behavior of our algorithm in terms of combined residual and cost function. We solve \eqref{eq:OurModel} for three choices of ${\mu}$ to show the sensitivity of our numerical algorithm with respect to ${\mu}$. Clearly seen, the numerical algorithm converges in a reasonable number of iterations for a wide range of ${\mu}$.

\section{Experiments}
We perform multiple experiments on benchmark image segmentation datasets to evaluate the performance of our method (termed {\PaperName}). The first part of this section gives information about the benchmarking datasets, evaluation measures, and parameter settings. The second part compares our results with state-of-the-art methods to demonstrate the effectiveness of {\PaperName}.

\subsection{Settings}
\noindent\textbf{Datasets:} We adopt three commonly used datasets in image segmentation: (1) BSD300 \cite{martin2001database} containing 300 images (200 training and 100 validation) of size $321\times 481$, where each image has in average 5 ground-truths manually drawn by human; (2) BSD500 \cite{arbelaez2011contour} is an extension of BSD300 with 200 new testing images; (3) MSRC \cite{shotton2009textonboost} containing 591 images of size $320\times 213$, where each image has a single ground-truth. It should be mentioned that we use the cleaned version of MSRC \cite{malisiewicz2007improving} in our experiments.

\noindent\textbf{Measures:} We adopt three widely accepted measures in image segmentation: (1) segmentation Covering (Cov) \cite{malisiewicz2007improving}, which measures the overlapping between two segmentations; (2) probability Rand Index (PRI) \cite{rand1971objective}, which measures the probability that a pair of pixels is consistently labeled in two segmentations; (3) variation of Information (VoI) \cite{meilǎ2005comparing}, which measures the distance between two segmentations as the average of their conditional entropy.

\noindent\textbf{Parameters:} Given an over-segmentation, {\PaperName} computes the superpixels features $\{{\mathbf{x}}_{\sm{i}{6}}\}_{\sm{i=1}{6}}^{\sm{n}{6}}$ by averaging the local spectral histogram (LSH) \cite{liu2006image} features of pixels within each superpixel. We use the algorithm and parameters presented in \cite{yuan2015factorization} to extract features and build a dictionary of size $l$. Parameter $l$ should be chosen sufficiently large (larger than the number of coherent regions) to ensure each superpixel feature is well expressible as a nonnegative linear combination of the dictionary words. We set $l=20$ which is normally much larger than the number of coherent regions in BSD and MSRC images. Our proposed model in \eqref{eq:OurModel} has two parameters $\gamma$ and $\lambda$, where $\gamma$ controls the effect of spatial relationship among superpixels and $\lambda$ controls the number of selected words. As $\gamma$ increases, the neighboring regions are more likely to be merged together and as $\lambda$ increases, the number of selected words reduces. We optimize $\gamma$ on the training set of BSD by applying grid search and use the optimized $\gamma$ in our experiments on BSD300, BSD500, and MSRC datasets. Parameter $\lambda$ is set to $\alpha {\lambda}_{\sm{\mathrm{max}}{6}}$, where $\alpha\!\in\!\left[0,1\right]$ and ${\lambda}_{\sm{\mathrm{max}}{6}}$ is a constant computed based on ${\mathbf{P}}\mathbf{R}^{\top}$, $\gamma$, and $\mathbf{L}$ using \cite{7364258}. If $\alpha$ is greater than $1$ (which means $\lambda\!>\!{\lambda}_{\sm{\mathrm{max}}{6}}$), only a single word is selected to represent the whole features. We follow \cite{arbelaez2011contour,Arbelaez_2014_CVPR,gao2016graph,ren2013image} to present our results as a family of segmentations which share the same parameter settings except for $\alpha$ that varies from $0$ to $1$. The evaluation measures are also reported at Optimal Dataset Scale (ODS) and Optimal Image Scale (OIS). 
\subsection{Results}
\noindent\textbf{Segmentation quality:} We run {\PaperName} on the benchmark datasets and report the results in tables \ref{tab:SegResBSD300}, \ref{tab:SegResBSD500}, and \ref{tab:SegResMSRC}, to make a comparison with recent methods such as, Normalized cut (Ncut) \cite{shi2000normalized}, Multi-scale Normalized cut (MNcut)\cite{cour2005spectral}, gPb-Ultametric contour map (gPb) \cite{arbelaez2011contour}, Image Segmentation by Cascade Region Agglomeration (ISCRA) \cite{ren2013image}, Reverse Image Segmentation with High/Low-level pairwise potentials (RIS-HL) \cite{wu2014reverse}, Multiscale Combinatorial Grouping (MCG) \cite{Arbelaez_2014_CVPR}, Contour-guided Color Palletes (CCP-2) \cite{fu2015robust}, Piecewise Flat Embedding (PFE) \cite{yu2015piecewise}, Discrete-Continuous Gradient Orientation Estimation for Segmentation (DC-Seg) \cite{donoser2014discrete}, Graph Without Cut (GWC) \cite{gao2016graph}, and Aligned hierarchical segmentation (MCG-Aligned) \cite{chen2016scale}. All scores are collected from \cite{arbelaez2011contour,Arbelaez_2014_CVPR,chen2016scale,donoser2014discrete,gao2016graph} except the MCG on MSRC and CCP-2 on BSD500 which are obtained by running the implementations provided by the respective authors.

\begin{table}
\centering
\scalebox{0.8}{%
\begin{tabular}{l||cc|cc|cc}
\toprule[\heavyrulewidth]
\multicolumn{1}{c}{} & \multicolumn{2}{c}{Cov (${\uparrow}$)}&  \multicolumn{2}{c}{PRI (${\uparrow}$)} &  \multicolumn{2}{c}{VoI (${\downarrow}$)} \\
\cmidrule(lr){2-3} \cmidrule(lr){4-5} \cmidrule(lr){6-7}

\multicolumn{1}{c}{Methods} & ODS & \multicolumn{1}{c}{OIS} & ODS & \multicolumn{1}{c}{OIS} & ODS & OIS\\
\midrule[\lightrulewidth]
MNcut\cite{cour2005spectral} &  $0.44$ & $0.53$ & $0.75$ & $0.79$ & $2.18$ & $1.84$\\[1mm]
gPb-UCM \cite{arbelaez2011contour} & $0.59$ & $0.65$ & $0.81$ & $0.85$ & $1.65$ & $1.47$\\[1mm]
ISCRA \cite{ren2013image}& $0.60$ & $0.67$ & $0.81$ & $\textbf{0.86}$ & $1.61$ &  $1.40$\\[1mm]
RIS+HL\cite{wu2014reverse}& $0.59$ & $0.65$ & $\textbf{0.82}$ & $\textbf{0.86}$ & $1.71$ & $1.53$\\[1mm]
MCG \cite{Arbelaez_2014_CVPR}& $\textbf{0.61}$ & $0.67$ & $0.81$ & $\textbf{0.86}$ & $1.55$ & $\textbf{1.37}$\\[1mm]
GWC \cite{gao2016graph}& $\textbf{0.61}$ & $\textbf{0.68}$ & $\textbf{0.82}$ & $\textbf{0.86}$ & $1.60$ & $1.42$\\[1mm]
{\PaperName}(MCG) & $\textbf{0.61}$ & $0.65$ & $0.81$ & $0.83$ & $\textbf{1.54}$ & $1.40$\\
\bottomrule
\end{tabular}}
   \caption{Quantitative comparisons on BSD300 val set.}
   \label{tab:SegResBSD300}
\end{table}
\begin{table}
\centering
\scalebox{0.8}{%
\begin{tabular}{l||cc|cc|cc}
\toprule[\heavyrulewidth]
\multicolumn{1}{c}{} & \multicolumn{2}{c}{Cov (${\uparrow}$)}&  \multicolumn{2}{c}{PRI (${\uparrow}$)} &  \multicolumn{2}{c}{VoI (${\downarrow}$)} \\
\cmidrule(lr){2-3} \cmidrule(lr){4-5} \cmidrule(lr){6-7}

\multicolumn{1}{c}{Methods} & ODS & \multicolumn{1}{c}{OIS} & ODS & \multicolumn{1}{c}{OIS} & ODS & OIS\\
\midrule[\lightrulewidth]
Ncut \cite{shi2000normalized} & $0.45$ & $0.53$ & $0.78$ & $0.80$ & $2.23$ & $1.89$\\[1mm]
gPb-UCM \cite{arbelaez2011contour} & $0.59$ & $0.65$ & $0.83$ & $0.86$ & $1.69$ & $1.48$\\[1mm]
DC-Seg \cite{donoser2014discrete} & $0.59$ & $0.64$ & $0.82$ & $0.85$ & $1.68$ & $1.54$\\[1mm] 
ISCRA \cite{ren2013image}& $0.59$ & $0.66$ & $0.82$ & $0.85$ & $1.60$ & $1.42$\\[1mm]
RIS+HL\cite{wu2014reverse}& $0.57$ & $0.66$ & $\textbf{0.84}$ & $0.86$ & $1.73$ & $1.55$\\[1mm]
MCG \cite{Arbelaez_2014_CVPR}& $0.61$ & $0.66$ & $0.83$ & $0.86$ & $1.57$ & $1.39$\\[1mm]
PFE+MCG \cite{yu2015piecewise}& $0.62$ & $\textbf{0.68}$ & $\textbf{0.84}$ & $\textbf{0.87}$ & $1.56$ & $1.36$\\[1mm]
MCG-Aligned \cite{chen2016scale}& $\textbf{0.63}$ & $\textbf{0.68}$ & $0.83$ & $0.86$ & $\textbf{1.53}$ & $1.38$\\[1mm]
GWC \cite{gao2016graph}& $0.61$ & $0.66$ & $0.83$ & $\textbf{0.87}$ & $1.62$ & $1.41$\\[1mm]
{\PaperName}(MCG)  & $\textbf{0.63}$ & $0.66$ & $0.83$ & $0.85$ & $1.55$ & $\textbf{1.35}$\\
\bottomrule
\end{tabular}}
   \caption{Quantitative comparisons on BSD500 test set.}
   \label{tab:SegResBSD500}
\end{table}
\begin{table}
\centering
\scalebox{0.8}{%
\begin{tabular}{l||cc|cc|cc}
\toprule[\heavyrulewidth]
\multicolumn{1}{c}{} & \multicolumn{2}{c}{Cov (${\uparrow}$)}&  \multicolumn{2}{c}{PRI (${\uparrow}$)} &  \multicolumn{2}{c}{VoI (${\downarrow}$)} \\
\cmidrule(lr){2-3} \cmidrule(lr){4-5} \cmidrule(lr){6-7}

\multicolumn{1}{c}{Methods} & ODS & \multicolumn{1}{c}{OIS} & ODS & \multicolumn{1}{c}{OIS} & ODS & OIS\\
\midrule[\lightrulewidth]
gPb-UCM \cite{arbelaez2011contour} & $0.65$ & $0.75$ & $0.78$ & $0.85$ & $1.28$ & $0.99$\\[1mm]
ISCRA \cite{ren2013image}& $0.67$ & $0.75$ & $0.77$ & $0.85$ & $1.18$ &  $1.02$\\[1mm]
GWC \cite{gao2016graph}& $0.68$ & $0.76$ & $0.78$ & $0.85$ & $1.24$ & $0.98$\\[1mm]
MCG \cite{Arbelaez_2014_CVPR}& $0.66$ &  $0.72$ & $0.78$ & $0.83$ & $1.23$ & $1.14$\\[1mm]
{\PaperName}(MCG)  & $\textbf{0.69}$ & $\textbf{0.77}$ & $\textbf{0.80}$ & $\textbf{0.86}$ & $\textbf{1.15}$ & $\textbf{0.91}$\\
\bottomrule
\end{tabular}}
   \caption{Quantitative comparisons on MSRC dataset.}
   \label{tab:SegResMSRC}
\end{table}
\newlength{\faribarulewidth}
\setlength{\faribarulewidth}{0.12mm}
\begin{table}
\centering
\scalebox{0.8}{%
\begin{tabular}{l||cc|cc|cc}
\toprule[\heavyrulewidth]
\multicolumn{1}{c}{} & \multicolumn{2}{c}{Cov (${\uparrow}$)}&  \multicolumn{2}{c}{PRI (${\uparrow}$)} &  \multicolumn{2}{c}{VoI (${\downarrow}$)} \\
\cmidrule(lr){2-3} \cmidrule(lr){4-5} \cmidrule(lr){6-7}

\multicolumn{1}{c}{Methods} & ODS & \multicolumn{1}{c}{OIS} & ODS & \multicolumn{1}{c}{OIS} & ODS & OIS\\
\midrule[\lightrulewidth]
ISCRA \cite{ren2013image}& $0.59$ & $0.66$ & $0.82$ & $0.85$ & $1.60$ & $1.42$\\[1mm]
{\PaperName} (ISCRA) \cite{comaniciu2002mean} & $0.61$ & $0.65$ & $0.82$ & $0.85$ & $1.58$ & $1.38$\\[1mm]
\midrule[\faribarulewidth]
CCP-2 \cite{fu2015robust} & $0.45$ & $-$ & $0.79$ & $-$ & $3.1$ & $-$\\[1mm]
{\PaperName} (CCP-2) & $0.58$ & $0.64$ & $0.81$ & $0.85$ & $1.78$ & $1.52$\\[1mm]
\midrule[\faribarulewidth]
MCG \cite{Arbelaez_2014_CVPR}& $0.61$ & $0.66$ & $0.83$ &  $0.86$ &  $1.57$ & $1.39$\\[1mm]
{\PaperName}(MCG)  & $0.63$ & $0.66$ & $0.83$ & $0.85$ & $1.55$ & $1.35$\\
\bottomrule
\end{tabular}}
   \caption{Sensitivity of our method with respect to the different initial over-segmentations on BSD500 test set.}
   \label{tab:SensOver}
\end{table}
\begin{table}
\centering
\scalebox{0.8}{%
\begin{tabular}{l||cc|cc|cc}
\toprule[\heavyrulewidth]
\multicolumn{1}{c}{} & \multicolumn{2}{c}{Cov (${\uparrow}$)}&  \multicolumn{2}{c}{PRI (${\uparrow}$)} &  \multicolumn{2}{c}{VoI (${\downarrow}$)} \\
\cmidrule(lr){2-3} \cmidrule(lr){4-5} \cmidrule(lr){6-7}

\multicolumn{1}{c}{$\gamma$} & ODS & \multicolumn{1}{c}{OIS} & ODS & \multicolumn{1}{c}{OIS} & ODS & OIS\\
\midrule[\lightrulewidth]
$10^{-2}$ & $0.62$ & $0.65$ & $0.83$ & $0.85$ & $1.56$ & $1.39$\\[1mm]
$10^{-1}$ & $0.62$ & $0.65$ & $0.83$ & $0.85$ & $1.55$ & $1.39$\\[1mm]
$10^{0}$ & $0.62$ & $0.66$ & $0.83$ & $0.85$ & $1.54$ & $1.38$\\[1mm]
$10^{1}$& $0.63$ & $0.66$ & $0.83$ & $0.85$ & $1.54$ & $1.37$\\[1mm]
$10^{2}$& $0.62$ & $0.64$ & $0.81$ &  $0.83$ &  $1.53$ & $1.41$\\[1mm]
\bottomrule
\end{tabular}}
   \caption{Sensitivity of our method with respect to the parameter variations on BSD500 test set.}
   \label{tab:SensParams}
\end{table}

\begin{figure*}
\captionsetup[subfigure]{position=b}
\centering
\subcaptionbox{\label{fig:finalResults1a}}{\begin{subfigure}[normal]{0.185\textwidth}
\centering
		\begin{subfigure}[normal]{0.411\linewidth}
			\includegraphics[width=\linewidth]{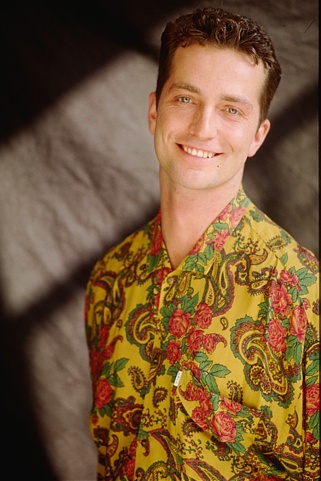}
		\end{subfigure}
		\begin{subfigure}[normal]{0.411\linewidth}
			\includegraphics[width=\linewidth]{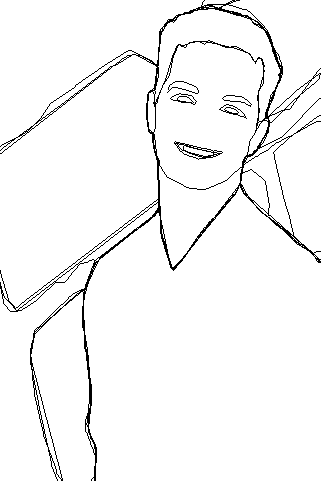}
		\end{subfigure}
        \\
		\begin{subfigure}[normal]{0.411\linewidth}
			\includegraphics[width=\linewidth]{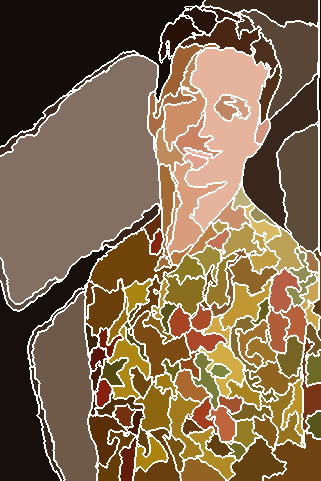}
		\end{subfigure}
		\begin{subfigure}[normal]{0.411\linewidth}
			\includegraphics[width=\linewidth]{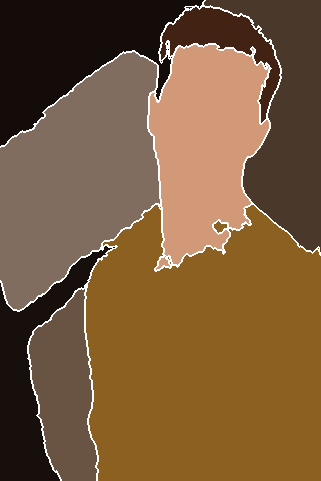}
		\end{subfigure}
		\\
		\begin{subfigure}[normal]{0.411\linewidth}
			\includegraphics[width=\linewidth]{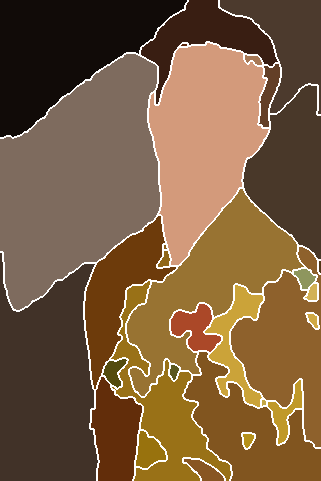}
		\end{subfigure}
		\begin{subfigure}[normal]{0.411\linewidth}
			\includegraphics[width=\linewidth]{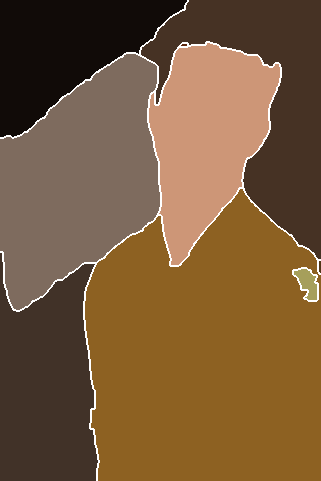}
		\end{subfigure}	
        \\
       \begin{subfigure}[normal]{0.411\linewidth}
			\includegraphics[width=\linewidth]{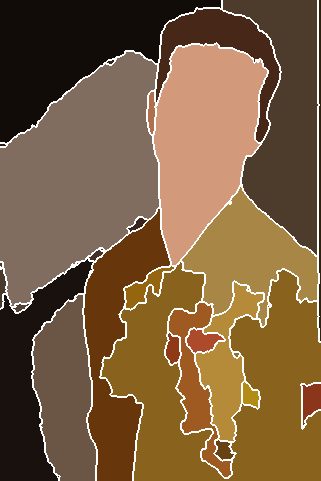}
		\end{subfigure}
		\begin{subfigure}[normal]{0.411\linewidth}
			\includegraphics[width=\linewidth]{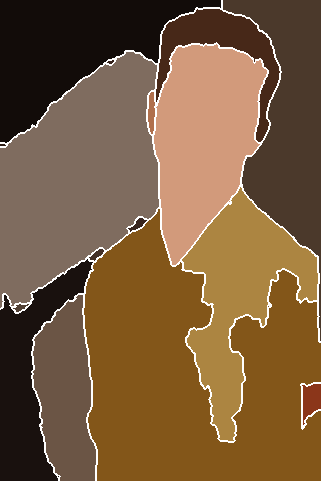}
		\end{subfigure}
\end{subfigure}}
\subcaptionbox{\label{fig:finalResults1b}}{\vspace{.35in}\begin{subfigure}[normal]{0.398\textwidth}
       	\begin{minipage}{0.192\linewidth}
        \centering
		\small{Image}
		\end{minipage}
        \begin{minipage}{0.192\linewidth}
        \centering
		\footnotesize{ISCRA}
		\end{minipage}
        \begin{minipage}{0.192\linewidth}
        \centering
		\footnotesize{MCG}
		\end{minipage}
        \begin{minipage}{0.192\linewidth}
        \centering
		\footnotesize{PFE+MCG}
		\end{minipage}
        \begin{minipage}{0.192\linewidth}
        \centering
		\footnotesize{{\PaperName}(MCG)}
		\end{minipage}
        \\       
		\begin{subfigure}[normal]{0.192\linewidth}
			\includegraphics[width=\linewidth]{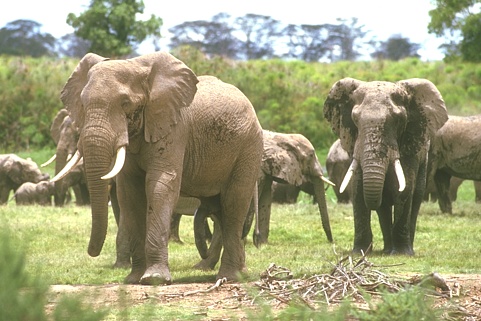}
		\end{subfigure}
		\begin{subfigure}[normal]{0.192\linewidth}
			\includegraphics[width=\linewidth]{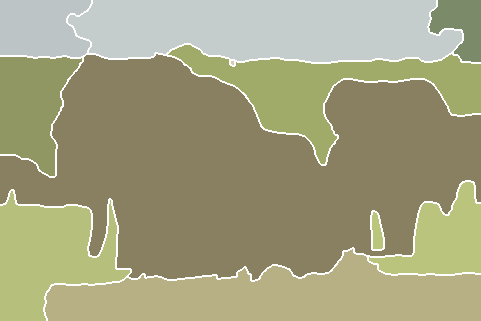}
		\end{subfigure}
		\begin{subfigure}[normal]{0.192\linewidth}
			\includegraphics[width=\linewidth]{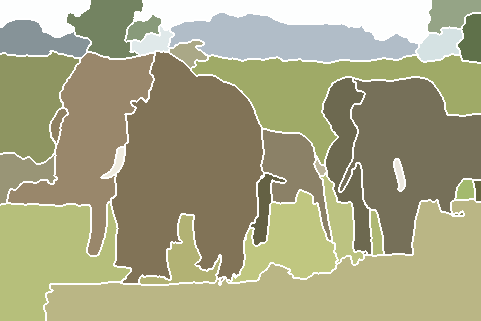}
		\end{subfigure}
		\begin{subfigure}[normal]{0.192\linewidth}
			\includegraphics[width=\linewidth]{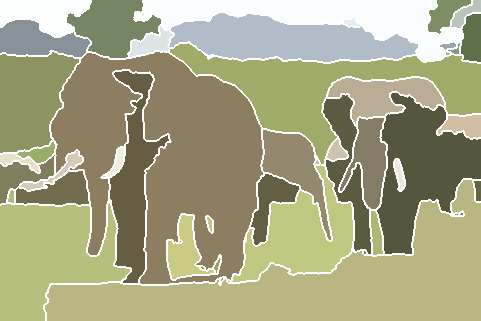}
		\end{subfigure}
		\begin{subfigure}[normal]{0.192\linewidth}
			\includegraphics[width=\linewidth]{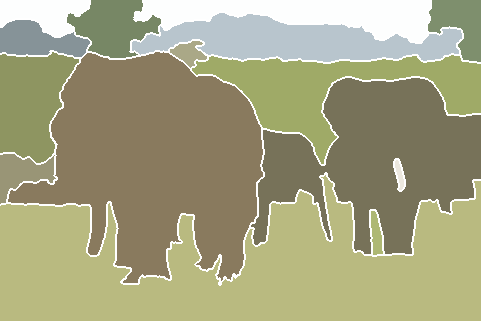}
		\end{subfigure}
        \\
		\begin{subfigure}[normal]{0.192\linewidth}
			\includegraphics[width=\linewidth]{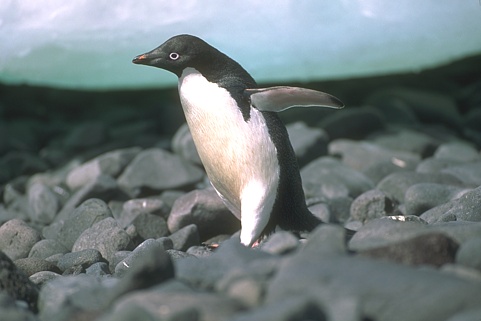}
		\end{subfigure}
		\begin{subfigure}[normal]{0.192\linewidth}
			\includegraphics[width=\linewidth]{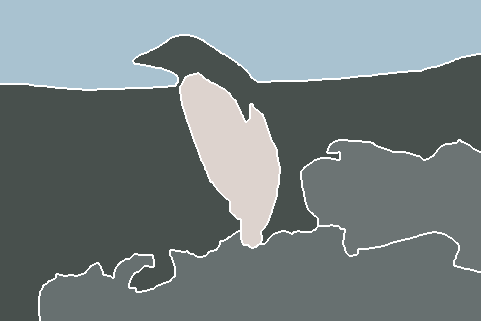}
		\end{subfigure}
		\begin{subfigure}[normal]{0.192\linewidth}
			\includegraphics[width=\linewidth]{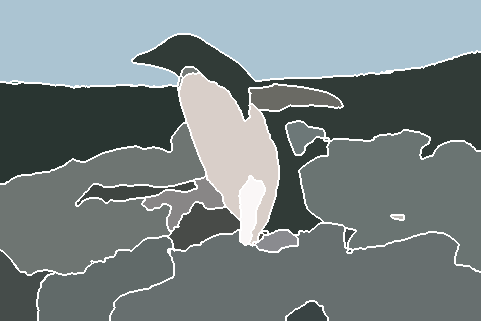}
		\end{subfigure}
		\begin{subfigure}[normal]{0.192\linewidth}
			\includegraphics[width=\linewidth]{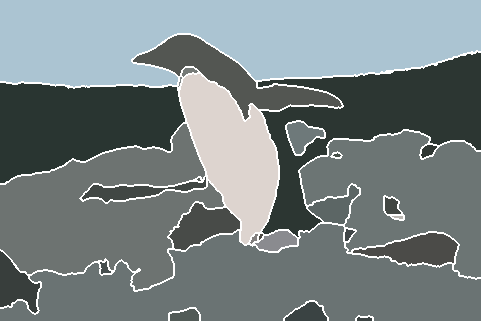}
		\end{subfigure}
		\begin{subfigure}[normal]{0.192\linewidth}
			\includegraphics[width=\linewidth]{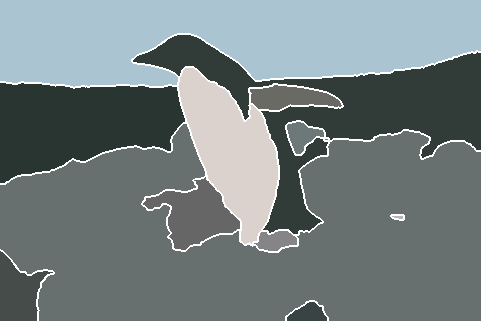}
		\end{subfigure}
		\\
		\begin{subfigure}[normal]{0.192\linewidth}
			\includegraphics[width=\textwidth]{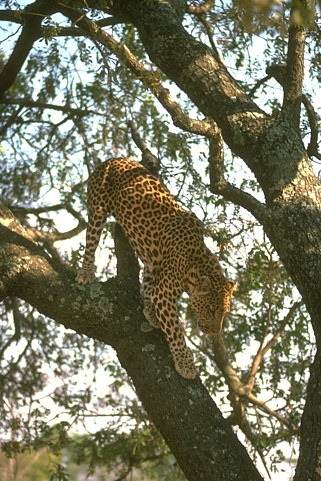}
		\end{subfigure}
		\begin{subfigure}[normal]{0.192\linewidth}
			\includegraphics[width=\linewidth]{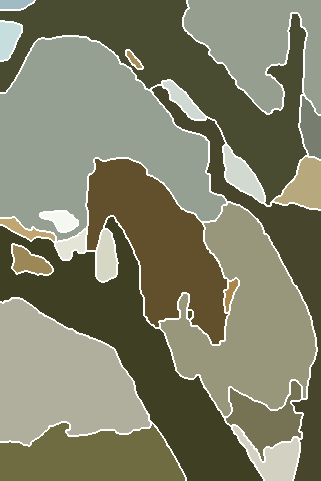}
		\end{subfigure}
		\begin{subfigure}[normal]{0.192\linewidth}
			\includegraphics[width=\linewidth]{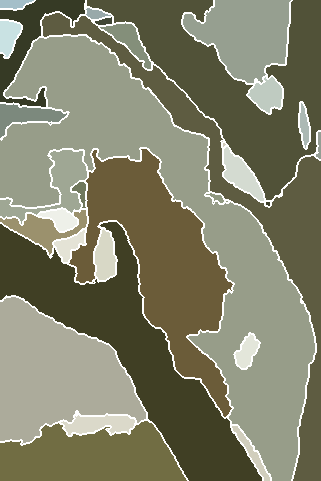}
		\end{subfigure}
		\begin{subfigure}[normal]{0.192\linewidth}
			\includegraphics[width=\linewidth]{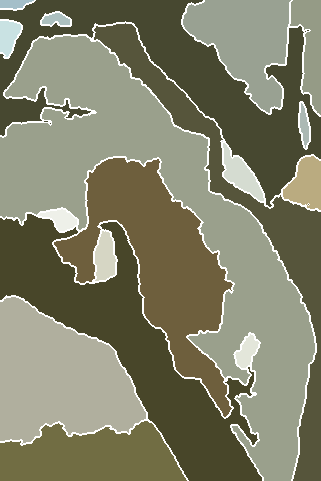}
		\end{subfigure}
		\begin{subfigure}[normal]{0.192\linewidth}
			\includegraphics[width=\linewidth]{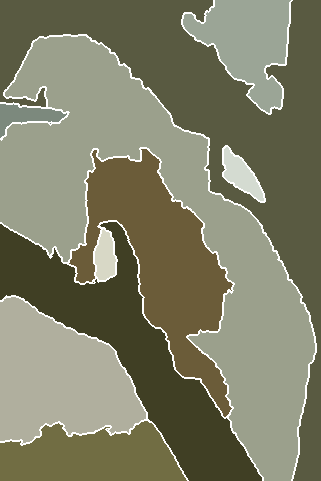}
		\end{subfigure}
		\\
		\begin{subfigure}[normal]{0.192\linewidth}
			\includegraphics[width=\linewidth]{Images/250047ori.jpg}
		\end{subfigure}
		\begin{subfigure}[normal]{0.192\linewidth}
			\includegraphics[width=\linewidth]{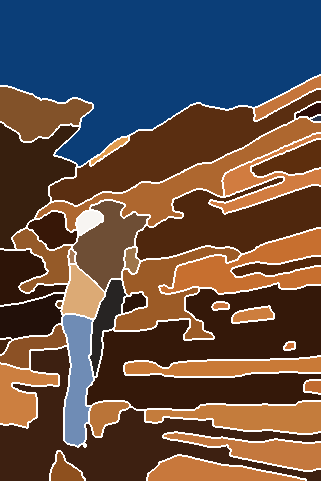}
		\end{subfigure}
		\begin{subfigure}[normal]{0.192\linewidth}
			\includegraphics[width=\linewidth]{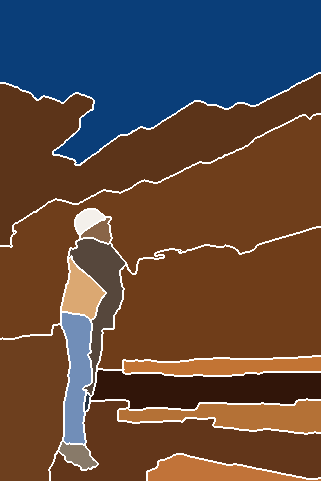}
		\end{subfigure}
		\begin{subfigure}[normal]{0.192\linewidth}
			\includegraphics[width=\linewidth]{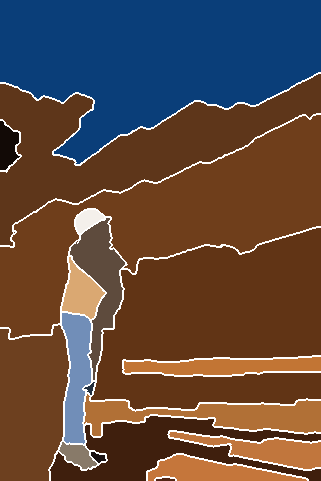}
		\end{subfigure}
		\begin{subfigure}[normal]{0.192\linewidth}
			\includegraphics[width=\linewidth]{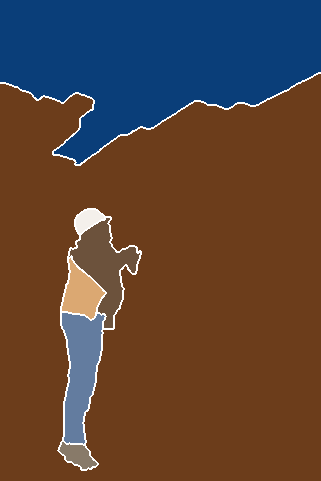}
		\end{subfigure}
\end{subfigure}
\hspace{0.01cm}
\begin{subfigure}[normal]{0.398\textwidth}
       	\begin{minipage}{0.192\linewidth}
        \centering
		\footnotesize{Image}
		\end{minipage}
        \begin{minipage}{0.192\linewidth}
        \centering
		\footnotesize{ISCRA}
		\end{minipage}
        \begin{minipage}{0.192\linewidth}
        \centering
		\footnotesize{MCG}
		\end{minipage}
        \begin{minipage}{0.192\linewidth}
        \centering
		\footnotesize{PFE+MCG}
		\end{minipage}
        \begin{minipage}{0.192\linewidth}
        \centering
		\footnotesize{{\PaperName}(MCG)}
		\end{minipage}
        \\       
		\begin{subfigure}[normal]{0.192\linewidth}
			\includegraphics[width=\linewidth]{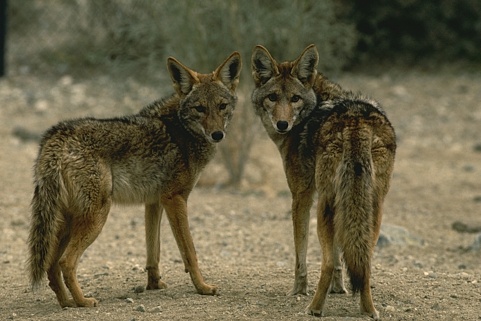}
		\end{subfigure}
		\begin{subfigure}[normal]{0.192\linewidth}
			\includegraphics[width=\linewidth]{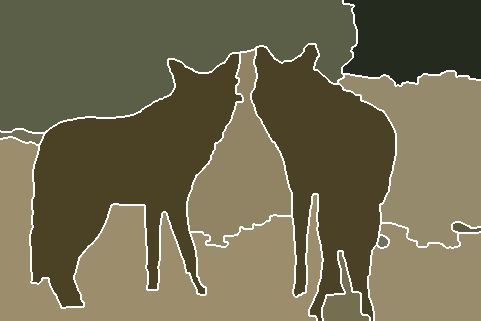}
		\end{subfigure}
		\begin{subfigure}[normal]{0.192\linewidth}
			\includegraphics[width=\linewidth]{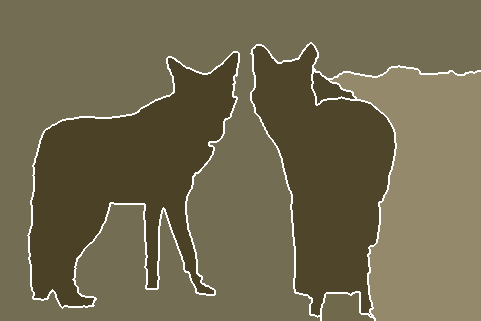}
		\end{subfigure}
		\begin{subfigure}[normal]{0.192\linewidth}
			\includegraphics[width=\linewidth]{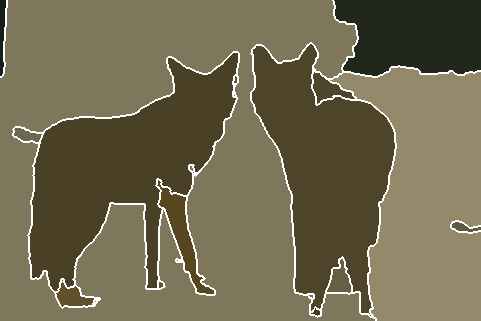}
		\end{subfigure}
		\begin{subfigure}[normal]{0.192\linewidth}
			\includegraphics[width=\linewidth]{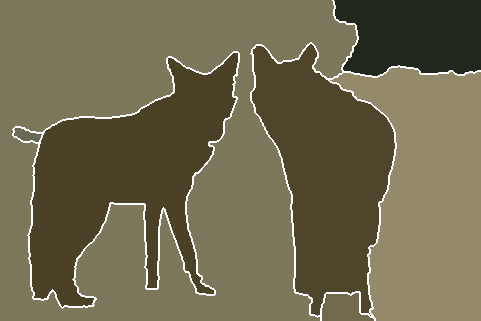}
		\end{subfigure}
        \\
        \begin{subfigure}[normal]{0.192\linewidth}
			\includegraphics[width=\linewidth]{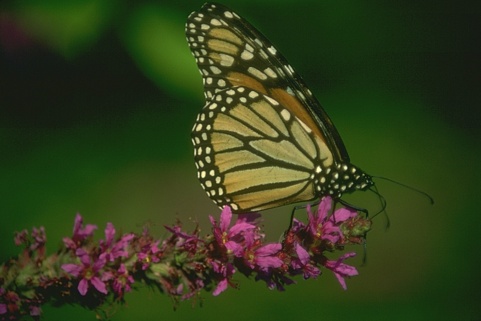}
		\end{subfigure}
		\begin{subfigure}[normal]{0.192\linewidth}
			\includegraphics[width=\linewidth]{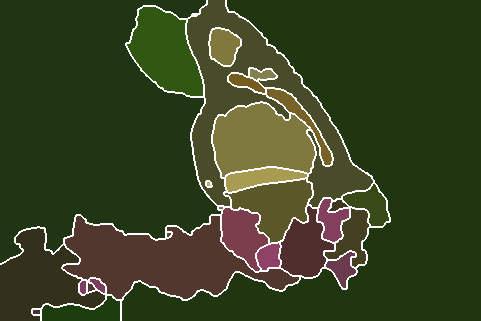}
		\end{subfigure}
		\begin{subfigure}[normal]{0.192\linewidth}
			\includegraphics[width=\linewidth]{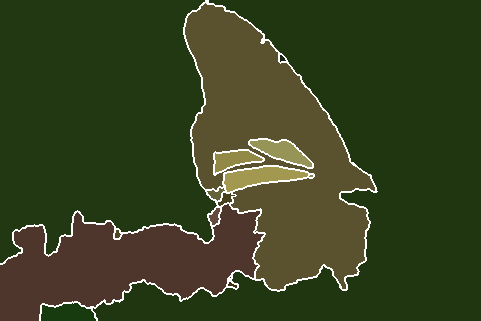}
		\end{subfigure}
		\begin{subfigure}[normal]{0.192\textwidth}
			\includegraphics[width=\linewidth]{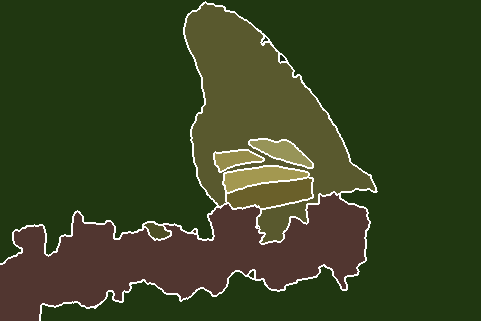}
		\end{subfigure}
		\begin{subfigure}[normal]{0.192\linewidth}
			\includegraphics[width=\linewidth]{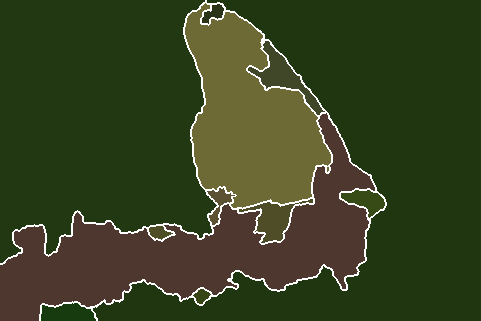}
		\end{subfigure}
		\\
		\begin{subfigure}[normal]{0.192\linewidth}
			\includegraphics[width=\textwidth]{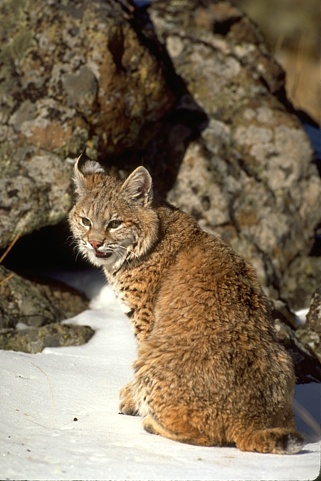}
		\end{subfigure}
		\begin{subfigure}[normal]{0.192\linewidth}
			\includegraphics[width=\linewidth]{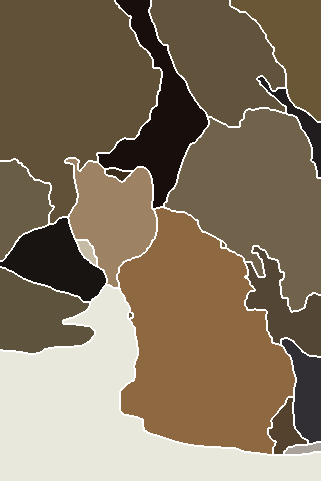}
		\end{subfigure}
		\begin{subfigure}[normal]{0.192\linewidth}
			\includegraphics[width=\linewidth]{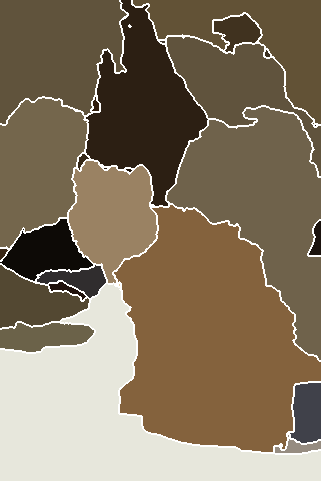}
		\end{subfigure}
		\begin{subfigure}[normal]{0.192\linewidth}
			\includegraphics[width=\linewidth]{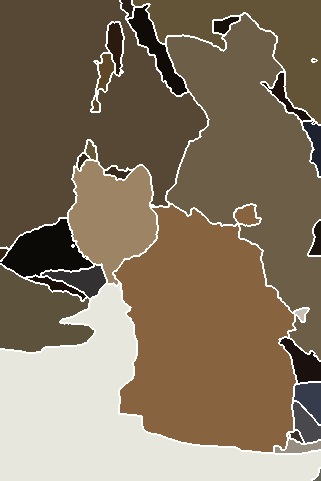}
		\end{subfigure}
		\begin{subfigure}[normal]{0.192\linewidth}
			\includegraphics[width=\linewidth]{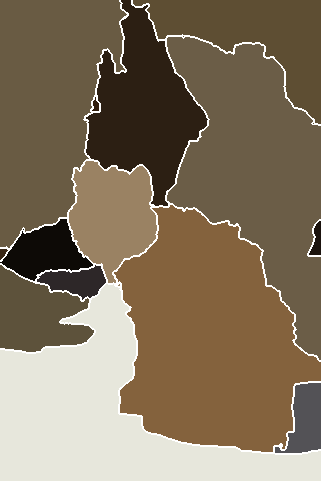}
		\end{subfigure}
		\\
		\begin{subfigure}[normal]{0.192\linewidth}
			\includegraphics[width=\linewidth]{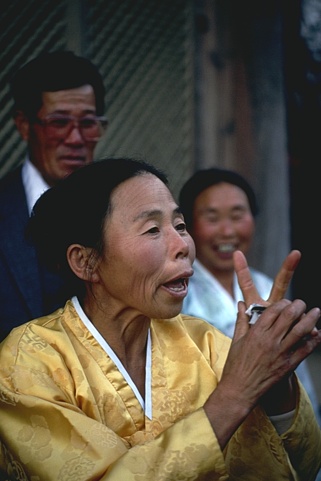}
		\end{subfigure}
		\begin{subfigure}[normal]{0.192\linewidth}
			\includegraphics[width=\linewidth]{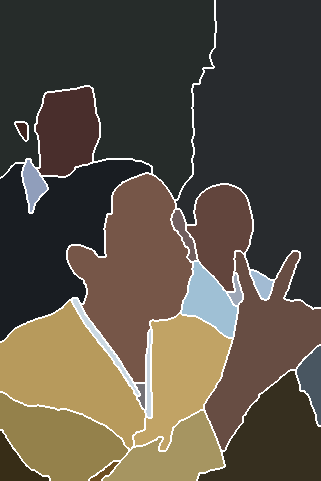}
		\end{subfigure}
		\begin{subfigure}[normal]{0.192\linewidth}
			\includegraphics[width=\linewidth]{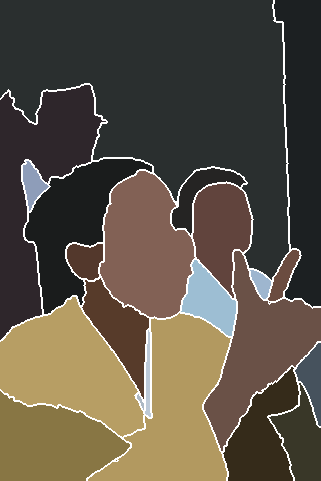}
		\end{subfigure}
		\begin{subfigure}[normal]{0.192\linewidth}
			\includegraphics[width=\linewidth]{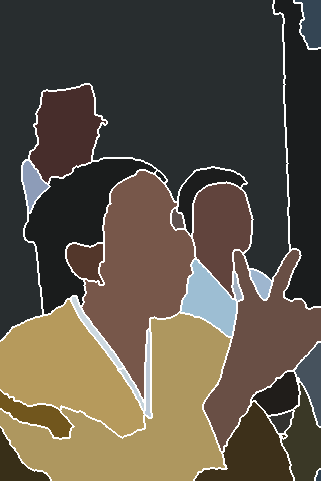}
		\end{subfigure}
		\begin{subfigure}[normal]{0.192\linewidth}
			\includegraphics[width=\linewidth]{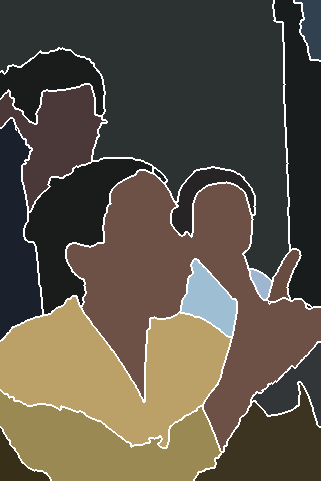}
		\end{subfigure}
\end{subfigure}}
\caption{Qualitative comparison of segmentations. (a) From top to bottom, left column: image, CCP-2 \cite{fu2015robust}, ISCRA \cite{ren2013image}, and MCG \cite{Arbelaez_2014_CVPR}; right column: groundtruth, {\PaperName}(CCP-2), {\PaperName}(ISCRA), and {\PaperName}(MCG); Note that the left column shows the initial over-segmentation provided by these methods and the right column shows our final segmentation with these initial over-segmentations. (b) All segmentation results are shown at Optimal Dataset Scale (ODS).}
\label{fig:finalResults1}
\end{figure*}

\noindent\textbf{Parameter sensitivity:} To evaluate the role played by an initial over-segmentation, we run {\PaperName} in combination with three segmentation methods CCP-2, ISRA, and MCG. In CCP-2, we use the same parameter settings as suggested by the respective author. In MCG and ISRA we respectively adopted the segmentations at scale $0.39$ and $44$ as the over-segmentations. In average, the over-segmentation layers provided by CCP-2, ISRA, and MCG have 120, 45, and 37 superpixels, respectively. Figure \ref{fig:finalResults1a} and table \ref{tab:SensOver} respectively show the qualitative and quantitative results of these combinations. Moreover, we run {\PaperName} for different $\gamma$ to assess our robustness with respect to the variations of $\gamma$. The results are reported in table \ref{tab:SensParams} in terms of segmentation measures. As tables \ref{tab:SensOver} and \ref{tab:SensParams} indicate, {\PaperName} not only achieves satisfactory results for a wide range of $\gamma$ but also improves the quality of initial over-segmentations on most of the segmentation measures. It is worth pointing out that {\PaperName} can be applied on the result of any segmentation method. The result may either be directly generated by a segmentation algorithm (e.g., CCP-2) or obtained from a specific level of a hierarchical segmentation (e.g., MCG).

Tables \ref{tab:SegResBSD300}, \ref{tab:SegResBSD500}, and \ref{tab:SegResMSRC} show that {\PaperName} in combination with MCG generates a high quality segmentation. The scores indicate that {\PaperName}(MCG) outperforms all competitor methods on BSD300 (ODS: VoI) and MSCRC (ODS: Cov, PRI, VoI and OIS: Cov, PRI, VoI). Other scores obtained by {\PaperName}(MCG) are also on par or in close proximity of the best competitors except for BSD300 (OIS: Cov, PRI) and BSD500 (OIS: PRI). In comparison with MCG, {\PaperName}(MCG) achieves better results on BSD300 (ODS: $0.01$ on VoI), BSD500 (ODS: $0.02$ on Cov, $0.02$ on VoI and OIS: $0.04$ on VoI), and MSRC (ODS: $0.03$ on Cov, $0.02$ on PRI, $0.08$ on VoI and OIS: $0.05$ on Cov, $0.03$ on PRI, $0.23$ on VoI). Our method is also on par with MCG on BSD300 (ODS: Cov, PRI) and BSD500 (ODS: PRI). Moreover, the tables indicate that {\PaperName}(MCG) achieves lower score than MCG in BSD300 (OIS: Cov, PRI, VoI) and BSD500 (OIS: PRI). It may seem reasonable to state that {\PaperName}(MCG) should consistently improve the MCG measures. However, this is not a fairly accurate statement. The reason is that {\PaperName} does not directly apply on the MCG hierarchical segmentation to improve or degrade the MCG results. It just adopts an over-segmentation by simply thresholding the MCG hierarchical segmentation and generates a family of segmentations. These segmentations differ from the ones which may be obtained at different thresholds of MCG.

MCG usually provides a high-quality hierarchical segmentation, but its performance is sometimes unsatisfactory, especially in textured images. Our method in combination with MCG improves the segmentation quality of these images (shown in Figure \ref{fig:finalResults1b}) by adopting superpixels features as an informative representation of small regions. Despite the advantages of {\PaperName}, it may fail to correctly segment the pixels belonging to an elongated coherent region. The reason is that the local neighborhood around these pixels contains visual information of neighboring coherent regions. Hence, their LSH features are inaccurate, which may cause a wrong assignment to the neighboring coherent regions.

\noindent\textbf{Algorithm complexity:} We decomposed model \eqref{eq:OurModel} into three sub-problems \eqref{eq:ADMMBlock1}, \eqref{eq:ADMMBlock2}, and \eqref{eq:ADMMBlock3} which are considered as the steps of the alternating direction method of multipliers (ADMM). To investigate the complexity of solving these sub-problems, we discuss each one separately. Subproblem \eqref{eq:ADMMBlock1} is cast as $l$ parallel equality constrained quadratic programs of form \eqref{eq:ADMMEq6} where each can be efficiently solved by solving a system of linear equations \cite{boyd2004convex} in $O(n^{\sm{2.8}{6}})$ \cite{strassen1969gaussian}. Since $\mathbf{A}$ and $\mathbf{B}$ are the same in all programs, the left-hand side of such systems are similar. Hence, one program just needs to be solved and then back-solves can be carried out for the right-hand sides of other programs. Therefore, the complexity of solving \eqref{eq:ADMMBlock1} is $O(n^{\sm{2.8}{6}})+O((l\!-\!1)n^{\sm{2}{6}})$. It is worth mentioning that in practice the complexity would be much lower, because $\mathbf{A}$ and $\mathbf{B}$ are highly sparse and structured. Subproblem \eqref{eq:ADMMBlock2} is split into two parallel problems \eqref{eq:ADMMEq7} and \eqref{eq:ADMMEq8}. Problem \eqref{eq:ADMMEq7} consists of $n$ parallel programs where each is solvable in $O(l\log(l))$ \cite{wang2013projection}. Problem \eqref{eq:ADMMEq8} consists of $n$ parallel programs where each is solved in $O(l)$ \cite{boyd2011distributed}. Therefore, \eqref{eq:ADMMBlock2} can be efficiently solved in $O(nl\log(l))$. Subproblem \eqref{eq:ADMMBlock3} consists of $n$ parallel updates over the columns of ${\mathbf{\Lambda}}_{\sm{1}{6}}$ and ${\mathbf{\Lambda}}_{\sm{2}{6}}$ which can be performed in $O(nl)$. In total, the complexity of our numerical solution is $O(n^{\sm{2.8}{6}})+O((l\!-\!1)n^{\sm{2}{6}})$ in the first iteration and $O(ln^{\sm{2}{6}})$ in subsequent iterations. Since all steps of our ADMM are highly parallelizable, the processing time can be significantly reduced using parallel computation. In the case of having $p$ parallel processing resources (assumption $n\!>\!l$), the complexity of \eqref{eq:ADMMBlock1}, \eqref{eq:ADMMBlock2}, and \eqref{eq:ADMMBlock3} are $O(n^{\sm{2.8}{6}})$, $O(\lceil\frac{n}{p}\rceil l\log(l))$, and $O(\lceil\frac{n}{p}\rceil l)$, respectively.

To compare our numerical algorithm with other solvers, we solve \eqref{eq:OurModel} for $20$ randomly chosen images in BSD500 using three standard convex solvers \cite{gb08,cvx}. The results are reported in tables \ref{tab:PerfCompa} and \ref{tab:PerfCompb} in terms of the average running times and relative errors, respectively. In this case, our relative error is computed as $\frac{{\left\|{\mathbf{U}}_{\sm{\mathrm{solver}}{4.5}}^{\sm{\ast}{4.5}}-{\mathbf{U}}_{\sm{\mathrm{\PaperName}}{4.5}}^{\sm{\ast}{5}}\right\|}_{\sm{F}{5}}}{{\left\|{\mathbf{U}}_{\sm{\mathrm{solver}}{5}}^{\sm{\ast}{5}}\right\|}_{\sm{F}{5}}}$. It should be noted that our average running time is computed without considering any parallelization.

\begin{table}
\begin{subtable}{\linewidth}
\centering
\scalebox{0.8}{%
\begin{tabular}{l||c|c|c|c}
\toprule[\heavyrulewidth]
\multicolumn{1}{c}{} & \multicolumn{1}{c}{SeDuMi} & \multicolumn{1}{c}{SDPT3} & \multicolumn{1}{c}{MOSEK} & \multicolumn{1}{c}{{\PaperName}(MCG)}\\
\midrule[\lightrulewidth ]
	\textbf{Run-Time}\;(sec.) & {$1.4\!\times\! 10^2$} & {$9.8\!\times\! 10^0$} & {$2.7\!\times\! 10^0$} & {$6.1\!\times\! 10^{-1}$} \\
\bottomrule
\end{tabular}}
\caption{}
\label{tab:PerfCompa}
\end{subtable}

\begin{subtable}{\linewidth}
\centering
\scalebox{0.8}{%
\begin{tabular}{l||c|c|c}
\toprule[\heavyrulewidth]
\multicolumn{1}{c}{} & \multicolumn{1}{c}{SeDuMi} & \multicolumn{1}{c}{SDPT3} & \multicolumn{1}{c}{MOSEK}\\
\midrule[\lightrulewidth ]
\textbf{Relative Error} & {$1.7\!\times\! 10^{-2}$} & {$1.1\!\times\!10^{-2}$} & {$9.0\!\times\! 10^{-3}$} \\

\bottomrule
\end{tabular}}
\caption{}
\label{tab:PerfCompb}
\end{subtable}
\caption{Performance comparison with convex solvers, SeDuMi \cite{sturm1999using}, SDPT3 \cite{toh1999sdpt3}, and MOSEK \cite{mosek2015mosek}.}
   \label{tab:PerfComp}
\end{table}

Table \ref{tab:PerfComp} indicates that our numerical algorithm is not only significantly faster than SeDuMi \cite{sturm1999using}, SDPT3 \cite{toh1999sdpt3}, and MOSEK \cite{mosek2015mosek} in solving \eqref{eq:OurModel}, but also offers an optimal solution extremely close to their solutions. The running time of {\PaperName} heavily depends on the initial over-segmentation. {\PaperName}(MCG) takes 5.8 seconds per image in average, 2.7 seconds for feature extraction, 2.5 seconds for learning the dictionary, and the remaining 0.6 seconds is taken by our numerical algorithm. All the experiments are performed on an Intel Core i5 quad-core 3.20GHz CPU and 16 GB RAM. 

\section{Conclusions}
This paper presented a novel segmentation model based on the concept of sparse subset selection. The model automatically estimates the optimal number of coherent regions and pixel assignments to form final segments. Moreover, we presented a parallel numerical algorithm based on the alternating direction method of multipliers (ADMM) to solve our model. The main advantages of this work are as follow: (1) does not require time for training over different datasets and works well in combination with various segmentation methods; (2) consists of three steps: extracting features, learning dictionary, and solving model where each one can be implemented in parallel; (3) contains ADMM steps with closed-form solutions which make the iterations computationally very efficient; (4) is not restricted to the segmentation problem and can be easily extended to other applications, such as video summarization, dimensionality reduction, etc.

{\small
\bibliographystyle{ieee}
\bibliography{egbib}

\begin{thebibliography}{10}\itemsep=-1pt

\bibitem{arbelaez2011contour}
P.~Arbelaez, M.~Maire, C.~Fowlkes, and J.~Malik.
\newblock Contour detection and hierarchical image segmentation.
\newblock {\em IEEE TPAMI}, 33(5):898--916, 2011.

\bibitem{Arbelaez_2014_CVPR}
P.~Arbelaez, J.~Pont-Tuset, J.~T. Barron, F.~Marques, and J.~Malik.
\newblock Multiscale combinatorial grouping.
\newblock In {\em CVPR}, 2014.

\bibitem{ben2001stability}
A.~Ben-Hur, A.~Elisseeff, and I.~Guyon.
\newblock A stability based method for discovering structure in clustered data.
\newblock In {\em PSB}, 2001.

\bibitem{boyd2011distributed}
S.~Boyd, N.~Parikh, E.~Chu, B.~Peleato, and J.~Eckstein.
\newblock Distributed optimization and statistical learning via the alternating
  direction method of multipliers.
\newblock {\em Foundations and Trends in Machine Learning}, 2011.

\bibitem{boyd2004convex}
S.~Boyd and L.~Vandenberghe.
\newblock {\em Convex optimization}.
\newblock Cambridge University Press, 2004.

\bibitem{chen2016scale}
Y.~Chen, D.~Dai, J.~Pont-Tuset, and L.~Van~Gool.
\newblock Scale-aware alignment of hierarchical image segmentation.
\newblock In {\em CVPR}, 2016.

\bibitem{comaniciu2002mean}
D.~Comaniciu and P.~Meer.
\newblock Mean shift: A robust approach toward feature space analysis.
\newblock {\em IEEE TPAMI}, 24(5):603--619, 2002.

\bibitem{cour2005spectral}
T.~Cour, F.~Benezit, and J.~Shi.
\newblock Spectral segmentation with multiscale graph decomposition.
\newblock In {\em CVPR}, 2005.

\bibitem{diaz2016lifting}
R.~D{\'\i}az, M.~Lee, J.~Schubert, and C.~C. Fowlkes.
\newblock Lifting gis maps into strong geometric context for scene
  understanding.
\newblock In {\em WACV}, 2016.

\bibitem{donoser2014discrete}
M.~Donoser and D.~Schmalstieg.
\newblock Discrete-continuous gradient orientation estimation for faster image
  segmentation.
\newblock In {\em CVPR}, 2014.

\bibitem{dutt2016active}
S.~Dutt~Jain and K.~Grauman.
\newblock Active image segmentation propagation.
\newblock In {\em CVPR}, 2016.

\bibitem{7364258}
E.~Elhamifar, G.~Sapiro, and S.~S. Sastry.
\newblock Dissimilarity-based sparse subset selection.
\newblock {\em IEEE TPAMI}, 38(11):2182--2197, Nov 2016.

\bibitem{elhamifar2012see}
E.~Elhamifar, G.~Sapiro, and R.~Vidal.
\newblock See all by looking at a few: Sparse modeling for finding
  representative objects.
\newblock In {\em CVPR}, 2012.

\bibitem{esser2012convex}
E.~Esser, M.~Moller, S.~Osher, G.~Sapiro, and J.~Xin.
\newblock A convex model for nonnegative matrix factorization and
  dimensionality reduction on physical space.
\newblock {\em IEEE TIP}, 21(7):3239--3252, 2012.

\bibitem{faridi2016automatic}
P.~Faridi, H.~Danyali, M.~S. Helfroush, and M.~A. Jahromi.
\newblock An automatic system for cell nuclei pleomorphism segmentation in
  histopathological images of breast cancer.
\newblock In {\em SPMB}, 2016.

\bibitem{felzenszwalb2004efficient}
P.~F. Felzenszwalb and D.~P. Huttenlocher.
\newblock Efficient graph-based image segmentation.
\newblock {\em IJCV}, 59(2):167--181, 2004.

\bibitem{fu2015robust}
X.~Fu, C.-Y. Wang, C.~Chen, C.~Wang, and C.-C. Jay~Kuo.
\newblock Robust image segmentation using contour-guided color palettes.
\newblock In {\em ICCV}, 2015.

\bibitem{gabay1976dual}
D.~Gabay and B.~Mercier.
\newblock A dual algorithm for the solution of nonlinear variational problems
  via finite element approximation.
\newblock {\em Computers \& Mathematics with Applications}, 2(1):17--40, 1976.

\bibitem{gao2016graph}
L.~Gao, J.~Song, F.~Nie, F.~Zou, N.~Sebe, and H.~T. Shen.
\newblock Graph-without-cut: An ideal graph learning for image segmentation.
\newblock In {\em AAAI}, 2016.

\bibitem{ghiasi2016laplacian}
G.~Ghiasi and C.~C. Fowlkes.
\newblock Laplacian pyramid reconstruction and refinement for semantic
  segmentation.
\newblock In {\em ECCV}, 2016.

\bibitem{goldstein2014fast}
T.~Goldstein, B.~O'Donoghue, S.~Setzer, and R.~Baraniuk.
\newblock Fast alternating direction optimization methods.
\newblock {\em SIAM Journal on Imaging Sciences}, 7(3):1588--1623, 2014.

\bibitem{gb08}
M.~Grant and S.~Boyd.
\newblock Graph implementations for nonsmooth convex programs.
\newblock In V.~Blondel, S.~Boyd, and H.~Kimura, editors, {\em Recent Advances
  in Learning and Control}, Lecture Notes in Control and Information Sciences,
  pages 95--110. Springer-Verlag Limited, 2008.

\bibitem{cvx}
M.~Grant and S.~Boyd.
\newblock {CVX}: Matlab software for disciplined convex programming, 2014.

\bibitem{hariharan2014simultaneous}
B.~Hariharan, P.~Arbel{\'a}ez, R.~Girshick, and J.~Malik.
\newblock Simultaneous detection and segmentation.
\newblock In {\em ECCV}, 2014.

\bibitem{jain2017fusionseg}
S.~D. Jain, B.~Xiong, and K.~Grauman.
\newblock Fusionseg: Learning to combine motion and appearance for fully
  automatic segmention of generic objects in videos.
\newblock {\em arXiv preprint arXiv:1701.05384}, 2017.

\bibitem{khoreva2014learning}
A.~Khoreva, F.~Galasso, M.~Hein, and B.~Schiele.
\newblock Learning must-link constraints for video segmentation based on
  spectral clustering.
\newblock In {\em GCPR}, 2014.

\bibitem{khoreva2015classifier}
A.~Khoreva, F.~Galasso, M.~Hein, and B.~Schiele.
\newblock Classifier based graph construction for video segmentation.
\newblock In {\em CVPR}, 2015.

\bibitem{kim2011boundary}
J.~Kim and K.~Grauman.
\newblock Boundary preserving dense local regions.
\newblock In {\em CVPR}, 2011.

\bibitem{kim2013learning}
T.~H. Kim, K.~M. Lee, and S.~U. Lee.
\newblock Learning full pairwise affinities for spectral segmentation.
\newblock {\em IEEE TPAMI}, 35(7):1690--1703, 2013.

\bibitem{kokkinos2015pushing}
I.~Kokkinos.
\newblock Pushing the boundaries of boundary detection using deep learning.
\newblock {\em arXiv preprint arXiv:1511.07386}, 2015.

\bibitem{krahenbuhl2011efficient}
P.~Kr{\"a}henb{\"u}hl and V.~Koltun.
\newblock Efficient inference in fully connected crfs with gaussian edge
  potentials.
\newblock In {\em NIPS}, 2011.

\bibitem{li2012segmentation}
Z.~Li, X.-M. Wu, and S.-F. Chang.
\newblock Segmentation using superpixels: A bipartite graph partitioning
  approach.
\newblock In {\em CVPR}, 2012.

\bibitem{7515198}
T.~Liu, M.~Seyedhosseini, and T.~Tasdizen.
\newblock Image segmentation using hierarchical merge tree.
\newblock {\em IEEE TIP}, 25(10):4596--4607, 2016.

\bibitem{liu2006image}
X.~Liu and D.~Wang.
\newblock Image and texture segmentation using local spectral histograms.
\newblock {\em IEEE TIP}, 15(10):3066--3077, 2006.

\bibitem{long2015fully}
J.~Long, E.~Shelhamer, and T.~Darrell.
\newblock Fully convolutional networks for semantic segmentation.
\newblock In {\em CVPR}, 2015.

\bibitem{malisiewicz2007improving}
T.~Malisiewicz and A.~A. Efros.
\newblock Improving spatial support for objects via multiple segmentations.
\newblock In {\em BMVC}, 2007.

\bibitem{Maninis2017}
K.~K. Maninis, J.~Pont-Tuset, P.~Arbeláez, and L.~Van~Gool.
\newblock Convolutional oriented boundaries: From image segmentation to
  high-level tasks.
\newblock {\em IEEE TPAMI}, PP(99):1--1, 2017.

\bibitem{martin2001database}
D.~Martin, C.~Fowlkes, D.~Tal, and J.~Malik.
\newblock A database of human segmented natural images and its application to
  evaluating segmentation algorithms and measuring ecological statistics.
\newblock In {\em ICCV}, 2001.

\bibitem{meilǎ2005comparing}
M.~Meilǎ.
\newblock Comparing clusterings: an axiomatic view.
\newblock In {\em ICML}, 2005.

\bibitem{mosek2015mosek}
A.~Mosek.
\newblock The mosek optimization toolbox for matlab manual.
\newblock {\em Version 7.1 (Revision 28)}, 2015.

\bibitem{noh2015learning}
H.~Noh, S.~Hong, and B.~Han.
\newblock Learning deconvolution network for semantic segmentation.
\newblock In {\em ICCV}, 2015.

\bibitem{rand1971objective}
W.~M. Rand.
\newblock Objective criteria for the evaluation of clustering methods.
\newblock {\em JASA}, 66(336):846--850, 1971.

\bibitem{ren2013image}
Z.~Ren and G.~Shakhnarovich.
\newblock Image segmentation by cascaded region agglomeration.
\newblock In {\em CVPR}, 2013.

\bibitem{shi2000normalized}
J.~Shi and J.~Malik.
\newblock Normalized cuts and image segmentation.
\newblock {\em IEEE TPAMI}, 22(8):888--905, 2000.

\bibitem{shotton2009textonboost}
J.~Shotton, J.~Winn, C.~Rother, and A.~Criminisi.
\newblock Textonboost for image understanding: Multi-class object recognition
  and segmentation by jointly modeling texture, layout, and context.
\newblock {\em IJCV}, 81(1):2--23, 2009.

\bibitem{still2004many}
S.~Still and W.~Bialek.
\newblock How many clusters? an information-theoretic perspective.
\newblock {\em Neural computation}, 16(12):2483--2506, 2004.

\bibitem{strassen1969gaussian}
V.~Strassen.
\newblock Gaussian elimination is not optimal.
\newblock {\em Numerische Mathematik}, 13(4):354--356, 1969.

\bibitem{sturm1999using}
J.~F. Sturm.
\newblock Using sedumi 1.02, a matlab toolbox for optimization over symmetric
  cones.
\newblock {\em Optimization methods and software}, 11(1-4):625--653, 1999.

\bibitem{tibshirani2001estimating}
R.~Tibshirani, G.~Walther, and T.~Hastie.
\newblock Estimating the number of clusters in a data set via the gap
  statistic.
\newblock {\em Journal of the Royal Statistical Society: Series B (Statistical
  Methodology)}, 63(2):411--423, 2001.

\bibitem{toh1999sdpt3}
K.-C. Toh, M.~J. Todd, and R.~H. T{\"u}t{\"u}nc{\"u}.
\newblock Sdpt3—a matlab software package for semidefinite programming,
  version 1.3.
\newblock {\em Optimization methods and software}, 11(1-4):545--581, 1999.

\bibitem{wang2013projection}
W.~Wang and M.~A. Carreira-Perpin{\'a}n.
\newblock Projection onto the probability simplex: An efficient algorithm with
  a simple proof, and an application.
\newblock {\em arXiv preprint arXiv:1309.1541}, 2013.

\bibitem{wu2014reverse}
J.~Wu, J.~Zhu, and Z.~Tu.
\newblock Reverse image segmentation: A high-level solution to a low-level
  task.
\newblock In {\em BMVC}, 2014.

\bibitem{yu2015piecewise}
Y.~Yu, C.~Fang, and Z.~Liao.
\newblock Piecewise flat embedding for image segmentation.
\newblock In {\em ICCV}, 2015.

\bibitem{yuan2015factorization}
J.~Yuan, D.~Wang, and A.~M. Cheriyadat.
\newblock Factorization-based texture segmentation.
\newblock {\em IEEE TIP}, 24(11):3488--3497, 2015.

\bibitem{zheng2015conditional}
S.~Zheng, S.~Jayasumana, B.~Romera-Paredes, V.~Vineet, Z.~Su, D.~Du, C.~Huang,
  and P.~H. Torr.
\newblock Conditional random fields as recurrent neural networks.
\newblock In {\em ICCV}, 2015.

\bibitem{zohrizadeh2016reliability}
F.~Zohrizadeh, M.~Kheirandishfard, K.~Ghasedidizaji, and F.~Kamangar.
\newblock Reliability-based local features aggregation for image segmentation.
\newblock In {\em ISVC}, 2016.

\end{thebibliography}
}

\end{document}